\documentclass{article} %
\usepackage[dvipsnames]{xcolor}
\usepackage{iclr2023_conference,times}

\usepackage{amsmath,amsfonts,bm}

\def\eqref#1{equation~\ref{#1}}

\def\1{\bm{1}}

\DeclareMathAlphabet{\mathsfit}{\encodingdefault}{\sfdefault}{m}{sl}
\SetMathAlphabet{\mathsfit}{bold}{\encodingdefault}{\sfdefault}{bx}{n}

\DeclareMathOperator*{\argmin}{arg\,min}

\newcommand{\Fig}[1]{\hyperref[{#1}]{Figure~\ref*{#1}}}  %
\newcommand{\Alg}[1]{\hyperref[{#1}]{Algorithm~\ref*{#1}}} 
\newcommand{\fig}[1]{\hyperref[{#1}]{Fig.~\ref*{#1}}}    %
\newcommand{\Tab}[1]{\hyperref[{#1}]{Table~\ref*{#1}}}
\newcommand{\tab}[1]{\hyperref[{#1}]{Table~\ref*{#1}}}
\newcommand{\Eqn}[1]{\hyperref[{#1}]{Equation~\ref*{#1}}}
\newcommand{\eqn}[1]{\hyperref[{#1}]{Eq.~\ref*{#1}}} %
\renewcommand{\sec}[1]{\hyperref[{#1}]{Sec.~\ref*{#1}}} %

\newcommand{\Sec}[1]{\hyperref[{#1}]{Section~\ref*{#1}}} %
\newcommand{\supp}[1]{\hyperref[{#1}]{Suppl.~\ref*{#1}}}
\newcommand{\app}[1]{\hyperref[{#1}]{App.~\ref*{#1}}}
\newcommand{\App}[1]{\hyperref[{#1}]{Appendix~\ref*{#1}}}

\usepackage{hyperref}
\usepackage{url}
\hypersetup{colorlinks,
            linkcolor=NavyBlue,
            urlcolor=NavyBlue,
            citecolor=NavyBlue}

\usepackage{array}
\newcolumntype{P}[1]{>{\centering\arraybackslash}p{#1}}
\usepackage{multirow}
\usepackage{booktabs}
\usepackage{amssymb}%
\usepackage{pifont}%
\newcommand{\xmark}{\ding{55}}%
\usepackage{makecell}
\usepackage{booktabs}
\usepackage{graphicx}
\usepackage{enumitem}
\usepackage{tablefootnote}
\definecolor{cyan}{cmyk}{.3,0,0,0}
\definecolor{lightblue}{HTML}{B0C4DE}
\definecolor{darkblue}{HTML}{177CB0}

\newcounter{rownumbers}
\usepackage{hyperref}
\usepackage{url}
\usepackage{bbm}
\usepackage{xcolor}
\usepackage{caption}

\definecolor{our-green}{rgb}{0.56, 0.692, 0.195}
\definecolor{our-darkgreen}{rgb}{0.297, 0.348, 0.105}
\definecolor{our-yellow}{rgb}{0.881, 0.611, 0.142}
\definecolor{our-red}{rgb}{0.923, 0.386, 0.209}

\usepackage[noend]{algpseudocode}
\algrenewcommand{\algorithmiccomment}[1]{%
\bgroup\hskip2em\textcolor{our-darkgreen}{//~\textsl{#1}}\egroup}

\usepackage{tabularx}
\usepackage{collcell}
\newcommand\setrow[1]{\gdef\rowmac{#1}\ignorespaces}
\newcommand\clearrow{\global\let\rowmac\relax}
\clearrow

\newcolumntype{C}{>{\collectcell\rowmac}c<{\endcollectcell}}
\newcolumntype{R}{>{\collectcell\rowmac}r<{\endcollectcell}}
\newcolumntype{L}{>{\collectcell\rowmac}l<{\endcollectcell}}

\usepackage{textcomp}
\usepackage{float}
\usepackage{wrapfig}
\usepackage[ruled]{algorithm2e}

\SetCommentSty{mycommfont}
\usepackage{adjustbox}
\newlength{\fHeight}
\newcommand{\faBicycle}{\includegraphics[height=\fHeight]{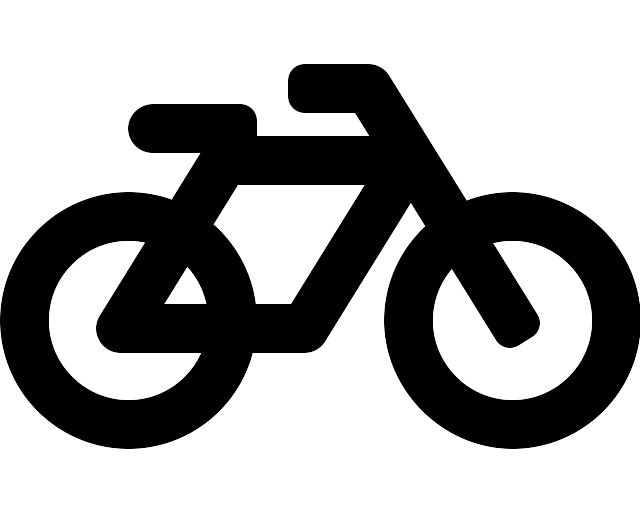}}
\newcommand{\faBird}{\includegraphics[height=\fHeight]{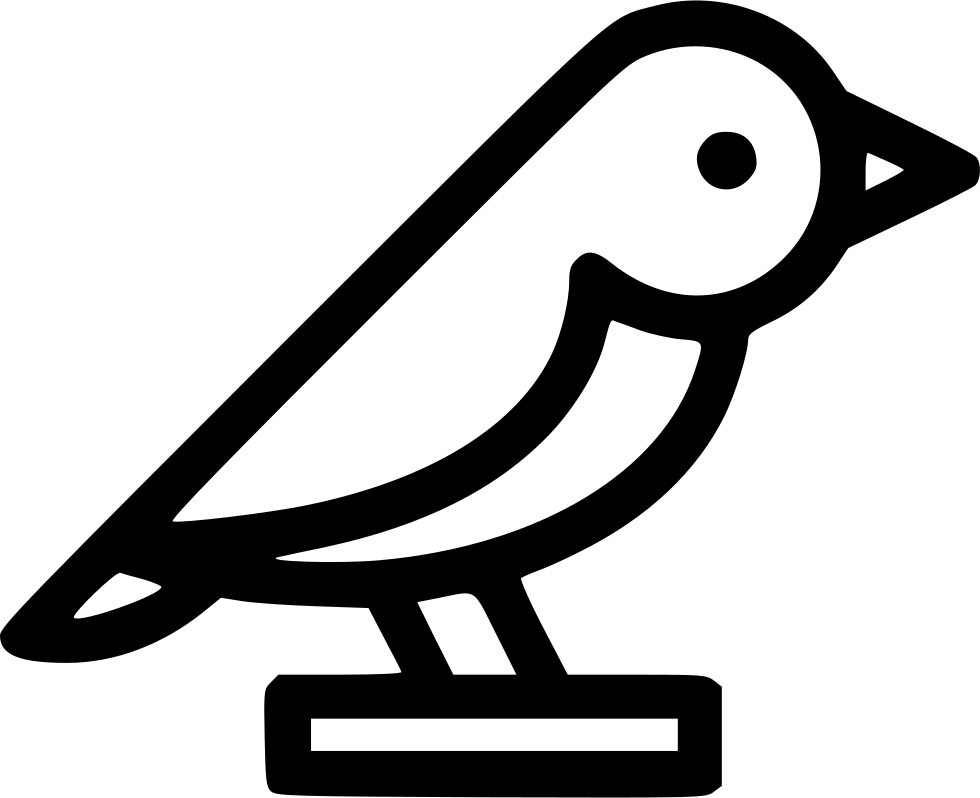}}
\newcommand{\faBus}{\includegraphics[height=\fHeight]{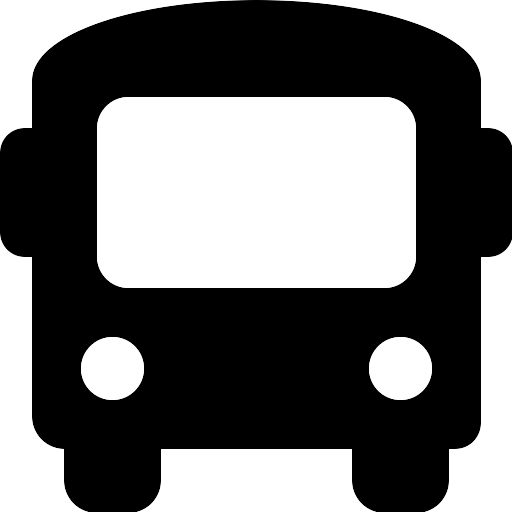}}
\newcommand{\faCar}{\includegraphics[height=\fHeight]{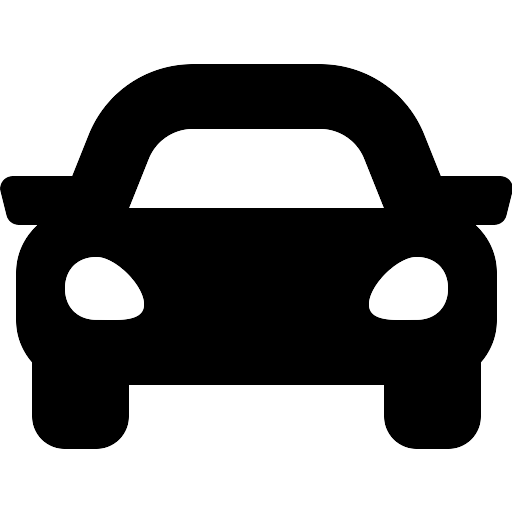}}
\newcommand{\faCat}{\includegraphics[height=\fHeight]{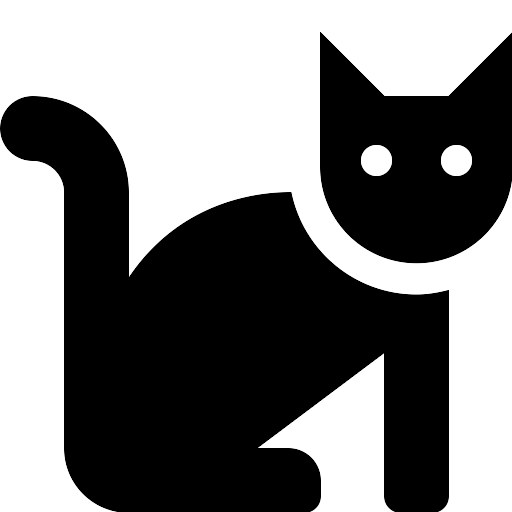}}
\newcommand{\faCow}{\includegraphics[height=\fHeight]{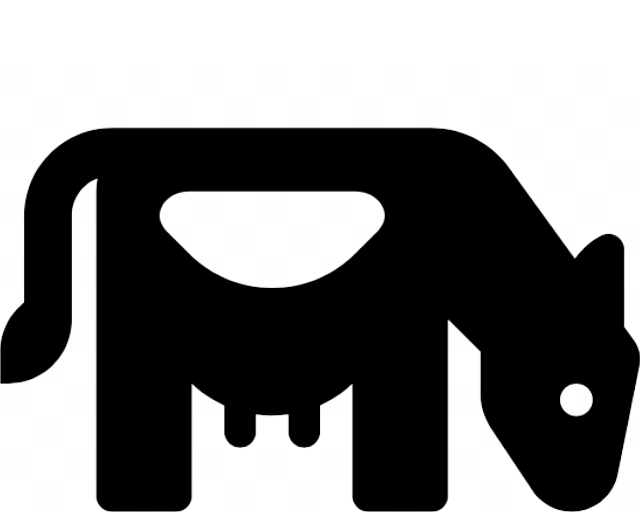}}
\newcommand{\faChair}{\includegraphics[height=\fHeight]{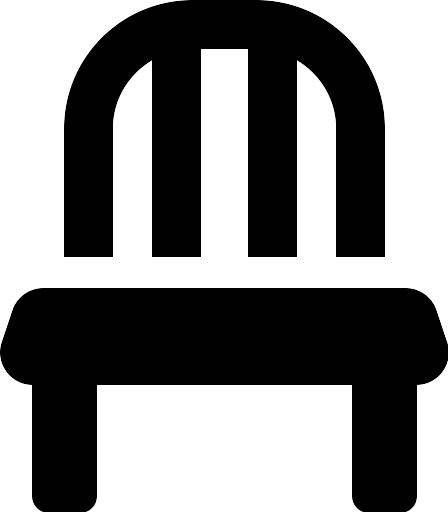}}
\newcommand{\faCouch}{\includegraphics[height=\fHeight]{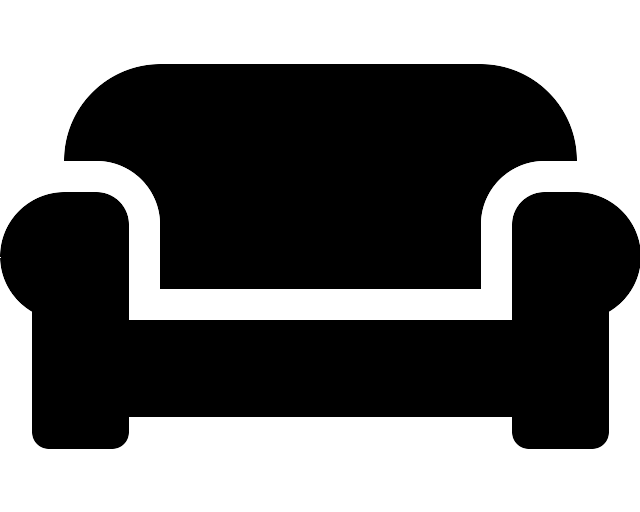}}

\newcommand{\faDog}{\includegraphics[height=\fHeight]{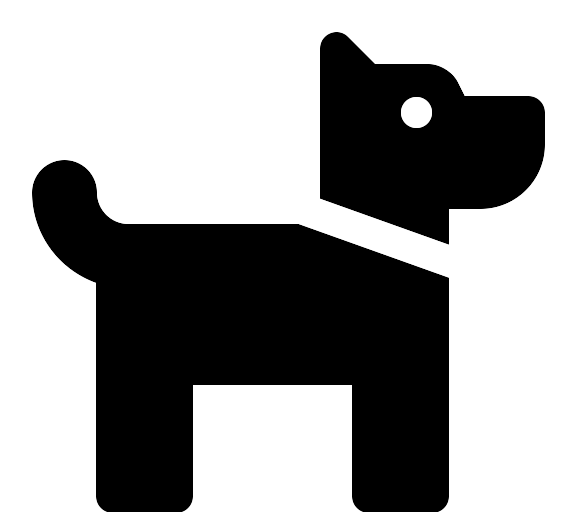}}
\newcommand{\faDinningTable}{\includegraphics[height=\fHeight]{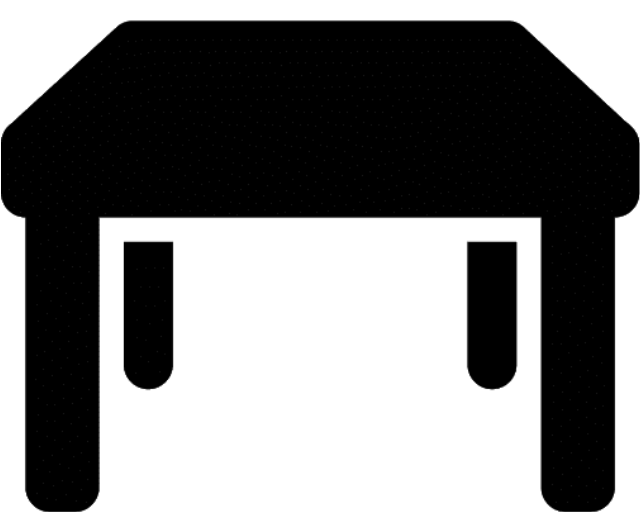}}
\newcommand{\faHorse}{\includegraphics[height=\fHeight]{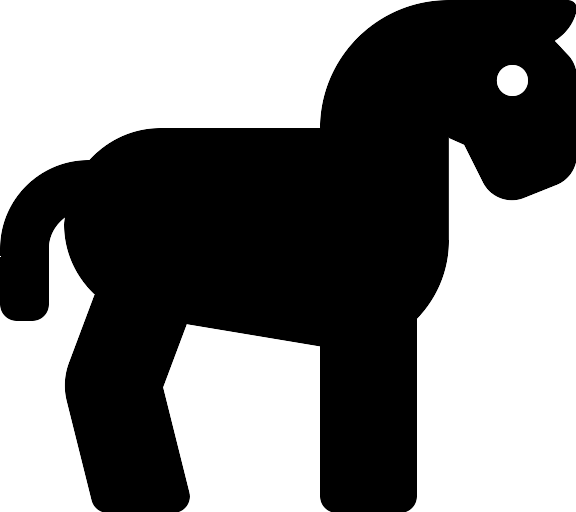}}
\newcommand{\faMotorcycle}{\includegraphics[height=\fHeight]{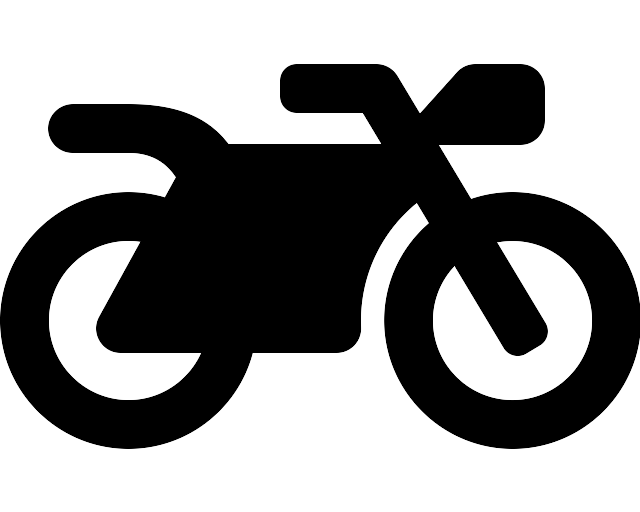}}
\newcommand{\faPlane}{\includegraphics[height=\fHeight]{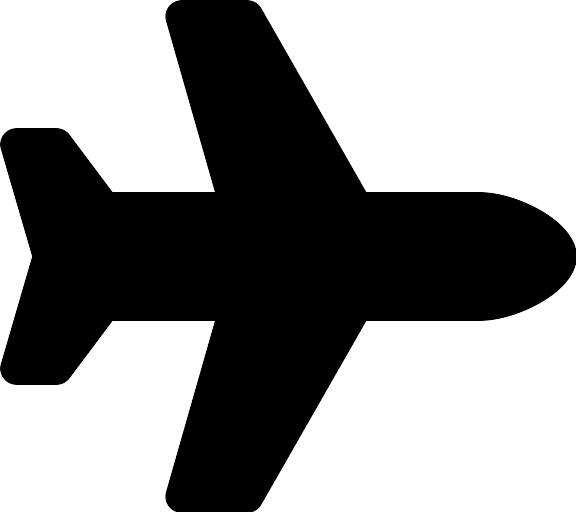}}
\newcommand{\faShip}{\includegraphics[height=\fHeight]{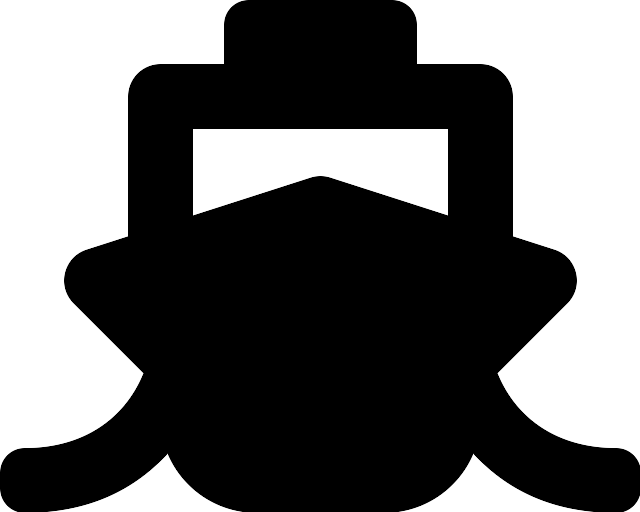}}
\newcommand{\faSheep}{\includegraphics[height=\fHeight]{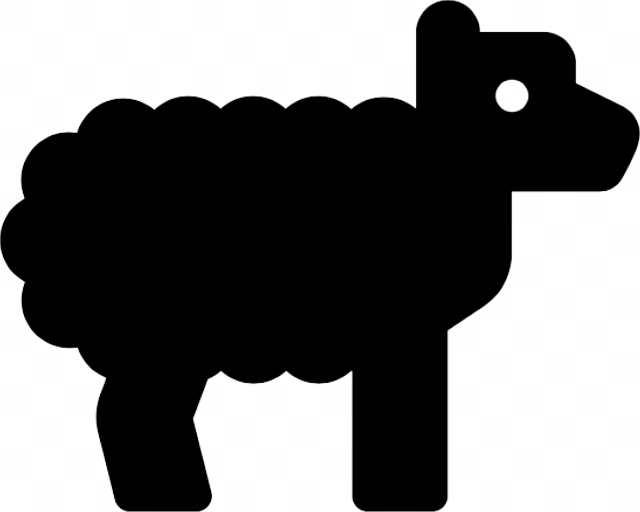}}
\newcommand{\faBackgraound}{\includegraphics[height=\fHeight]{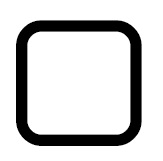}}
\newcommand{\faTrain}{\includegraphics[height=\fHeight]{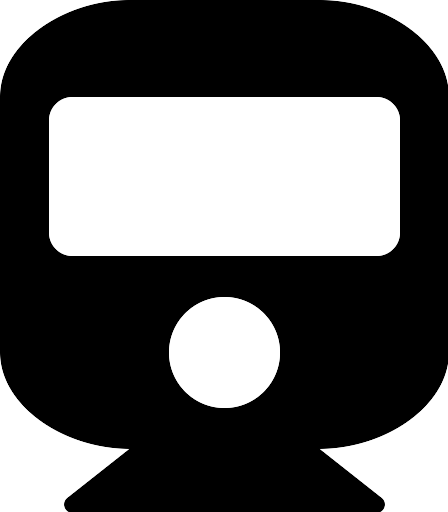}}
\newcommand{\faTulip}{\includegraphics[height=\fHeight]{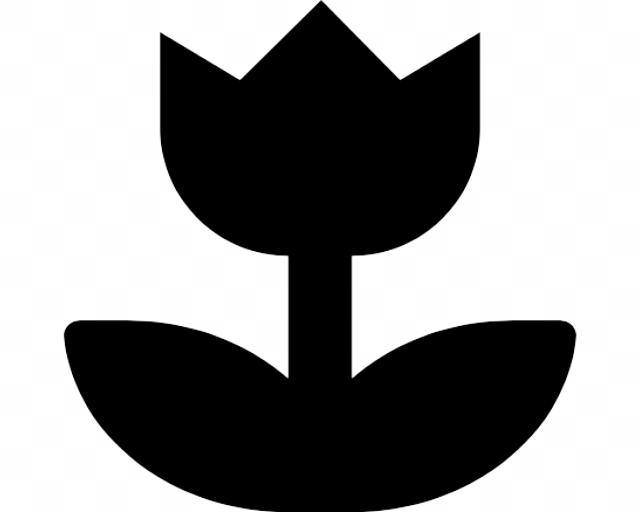}}
\newcommand{\faTv}{\includegraphics[height=\fHeight]{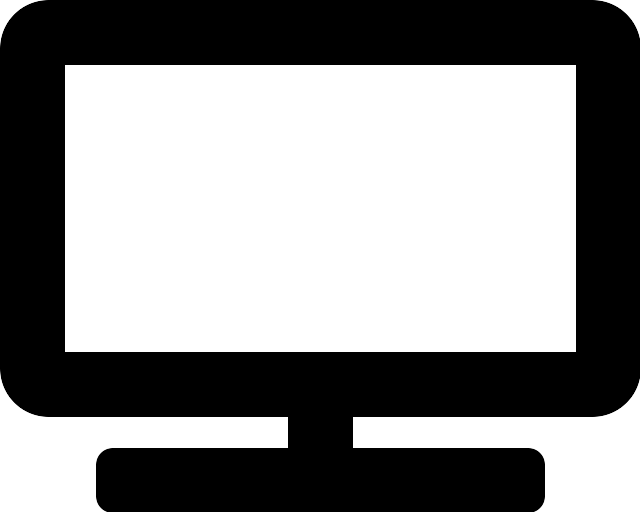}}
\newcommand{\faWalking}{\includegraphics[height=\fHeight]{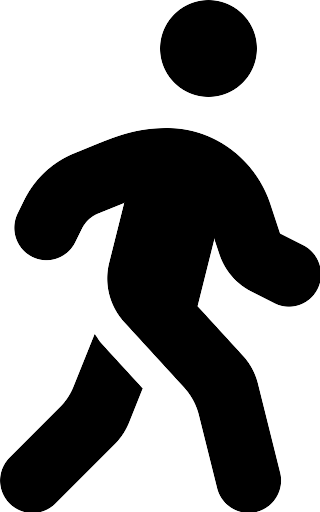}}
\newcommand{\faWineBottle}{\includegraphics[height=\fHeight]{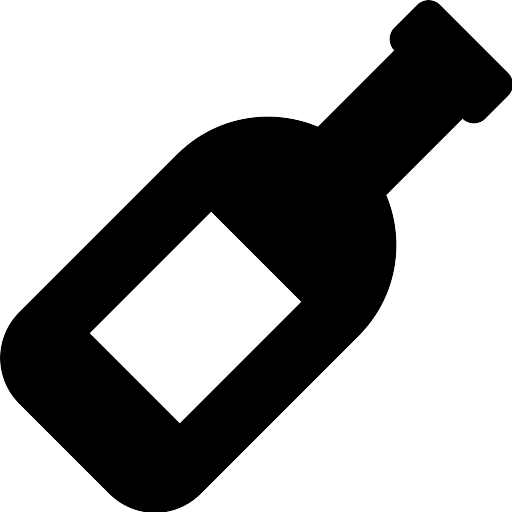}}

\DeclareMathOperator{\crop}{crop}
\footnotesize
\newcommand{\tableIcons}{%
\faBackgraound & \faPlane & \faBicycle & \faBird & \faShip & \faWineBottle & \faBus & \faCar & \faCat &	\faChair & \faCow & \faDinningTable & \faDog & \faHorse & \faMotorcycle & \faWalking & \faTulip & \faSheep & \faCouch & \faTrain & \faTv}

\newcommand{\method}{COMUS\xspace}

\title{Unsupervised Semantic Segmentation with Self-supervised Object-centric Representations}

\author{Andrii Zadaianchuk\textsuperscript{1,3}\thanks{Work done during internship at Amazon.} ,
Matthaeus Kleindessner\textsuperscript{2},
Yi Zhu\textsuperscript{2},
\textbf{Francesco Locatello}\textsuperscript{2},
\textbf{Thomas Brox}\textsuperscript{2,4} \\
\textsuperscript{1}Max-Planck Institute for Intelligent Systems, T\"ubingen, Germany,
\textsuperscript{2}Amazon Web Services, \\
\textsuperscript{3}Department of Computer Science, ETH Z\"urich,
\textsuperscript{4}University of Freiburg \\
}

\iclrfinalcopy %
\begin{document}

\maketitle

\begin{abstract}
In this paper, we show that recent advances in self-supervised representation learning enable \textit{unsupervised} object discovery and semantic segmentation with a performance that matches the state of the field on \textit{supervised} semantic segmentation 10 years ago. We propose a methodology based on unsupervised saliency masks and self-supervised feature clustering to kickstart object discovery followed by training a semantic segmentation network on pseudo-labels to bootstrap the system on images with multiple objects. We show that while being conceptually simple our proposed baseline is surprisingly strong. We present results on PASCAL VOC that go far beyond the current state of the art (50.0 mIoU)
, and we report for the first time results on MS COCO for the whole set of 81 classes: our method discovers 34 categories with more than $20\%$ IoU, while obtaining an average IoU of 19.6 for all 81 categories.
\end{abstract}
\begin{figure}[ht]
    
   \includegraphics[width=\linewidth]{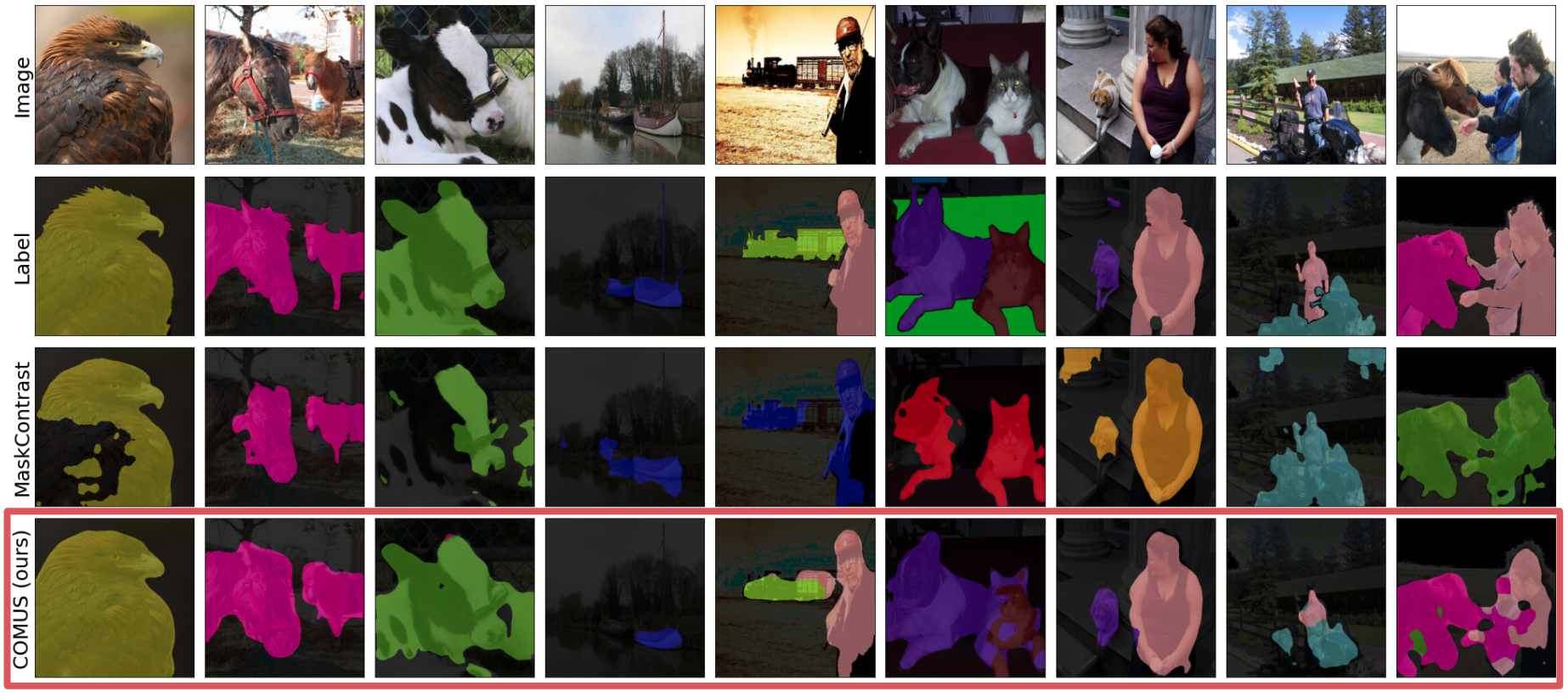}
   \caption{Unsupervised semantic segmentation predictions on PASCAL VOC~\citep{pascal-voc-2012}. Our \method does not use human annotations to discover objects and their precise localization. In contrast to the prior state-of-the-art method MaskContrast~\citep{vangansbeke2020unsupervised},
  \method yields more precise segmentations, avoids confusion of categories, and is not restricted to only one object category per image.}
   \label{fig:vis}
\end{figure}

\section{Introduction}

The large advances in dense semantic labelling in recent years were built on large-scale human-annotated datasets~\citep{pascal-voc-2012, Lin2014MicrosoftCC,cordts2016cityscapes}. These supervised semantic segmentation methods~\citep[e.g.,][]{Ronneberger2015UNetCN, Chen2018DeepLabSI} require costly human annotations and operate only on a restricted set of predefined categories.  
Weakly-supervised segmentation~\citep{pathakICLR15, Wei_2018_CVPR} and semi-supervised segmentation~\citep{MTB19, Zhu2020ImprovingSS} approach the issue of annotation cost by reducing the annotation to only a class label or to a subset of labeled images. However, they are still bound to predefined labels. 

In this paper, we follow a recent trend to move away from the external definition of class labels and rather try to identify object categories automatically by letting the patterns in the data speak. This could be achieved by (1) exploiting dataset biases to replace the missing annotation, (2) a way to get the learning process kickstarted based on ``good'' samples, and (3) a bootstrapping process that iteratively expands the domain of exploitable samples. 

\begin{figure}[t]
   \includegraphics[width=\linewidth]{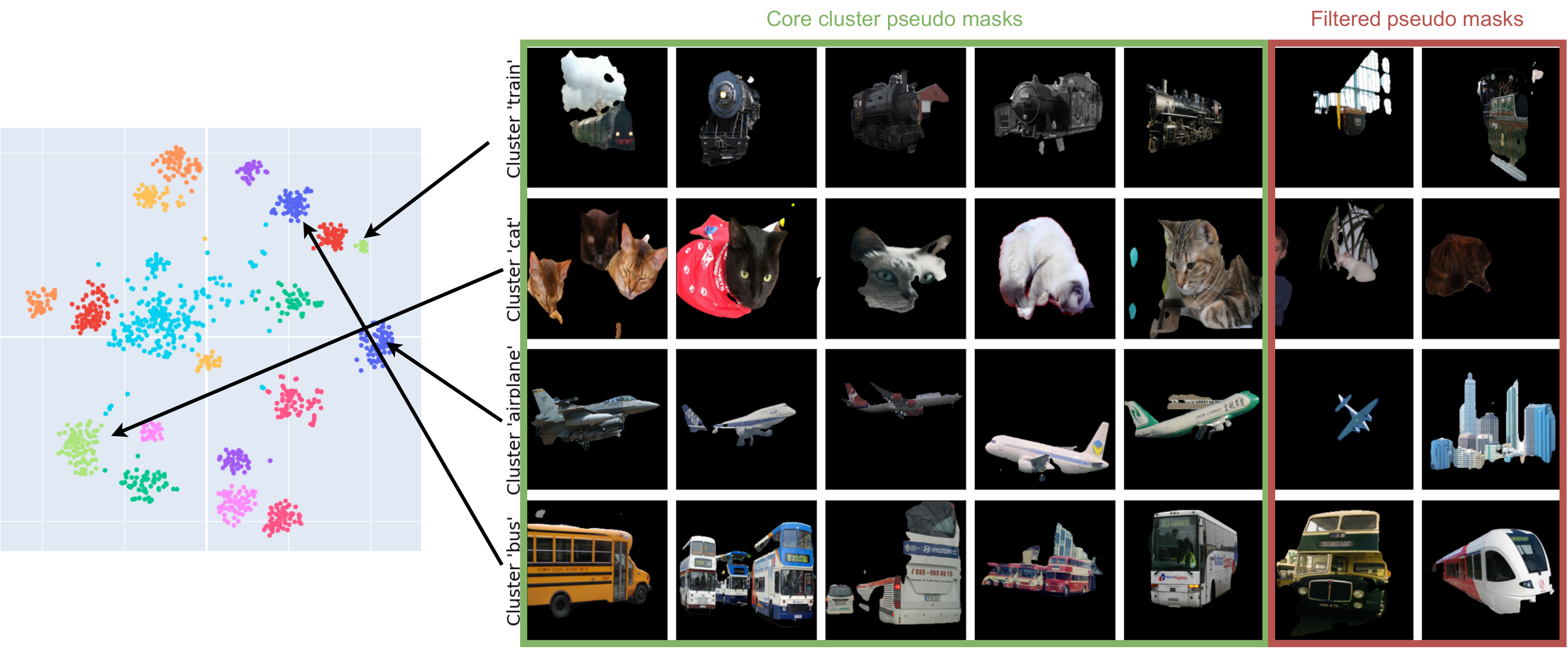}
   \caption{Visualization of unsupervised pseudo-masks on PASCAL VOC \textit{val} set. \textit{(left)} 2D t-SNE projection of object proposal features. Colors correspond to cluster IDs. \textit{(right)} Pseudo-masks from different clusters. The pseudo-masks were randomly sampled for each cluster from both cluster core pseudo-masks (green columns) and filtered pseudo-masks~(red columns).}
   \label{fig:vis_clustering}
\end{figure}

A recent method that exploits dataset biases, DINO~\citep{caronEmergingPropertiesSelfSupervised2021}, reported promising effects of self-supervised feature learning in conjunction with a visual transformer architecture by exploiting the object-centric bias of ImageNet with a multi-crop strategy. Their paper emphasized particularly the object-centric attention maps on some samples. We found that the attention maps of their DINO approach are not strong enough on a broad enough set of images to kickstart unsupervised semantic segmentation~(see \fig{fig:vis_masks}), but their learned features within an object region yield clusters of surprisingly high purity and align well with underlying object categories~(see \fig{fig:vis_clustering}). 

Thus, we leverage unsupervised saliency maps from DeepUSPS~\citep{Nguyen_Neurips2019} and BASNet~\citep{Qin_2019_CVPR} to localize foreground objects and to extract DINO features from these foreground regions. 
This already enables unsupervised semantic segmentation on images that show a dominant object category together with an unspectacular background as they are common in PASCAL VOC~\citep{pascal-voc-2012}. 
However, on other datasets, such as MS COCO~\citep{Lin2014MicrosoftCC}, most objects are in context with other objects. Even on PASCAL VOC, there are many images with multiple different object categories. 

For extending to more objects, we propose training a regular semantic segmentation network on the obtained pseudo-masks and to further refine this network by self-training it on its own outputs. Our method, dubbed \textbf{\method} (for \textbf{C}lustering \textbf{O}bject \textbf{M}asks for learning \textbf{U}nsupervised \textbf{S}egmentation), allows us to segment objects also in multi-object images (see \Fig{fig:vis}), and it allows us for the first time to report unsupervised semantic segmentation results on the full 80 category MS COCO dataset without any human annotations. While there are some hard object categories that are not discovered by our proposed procedure, we obtain good clusters for many of COCO object categories.\\[-1mm]

\begin{figure}[t]
   \includegraphics[width=\linewidth]{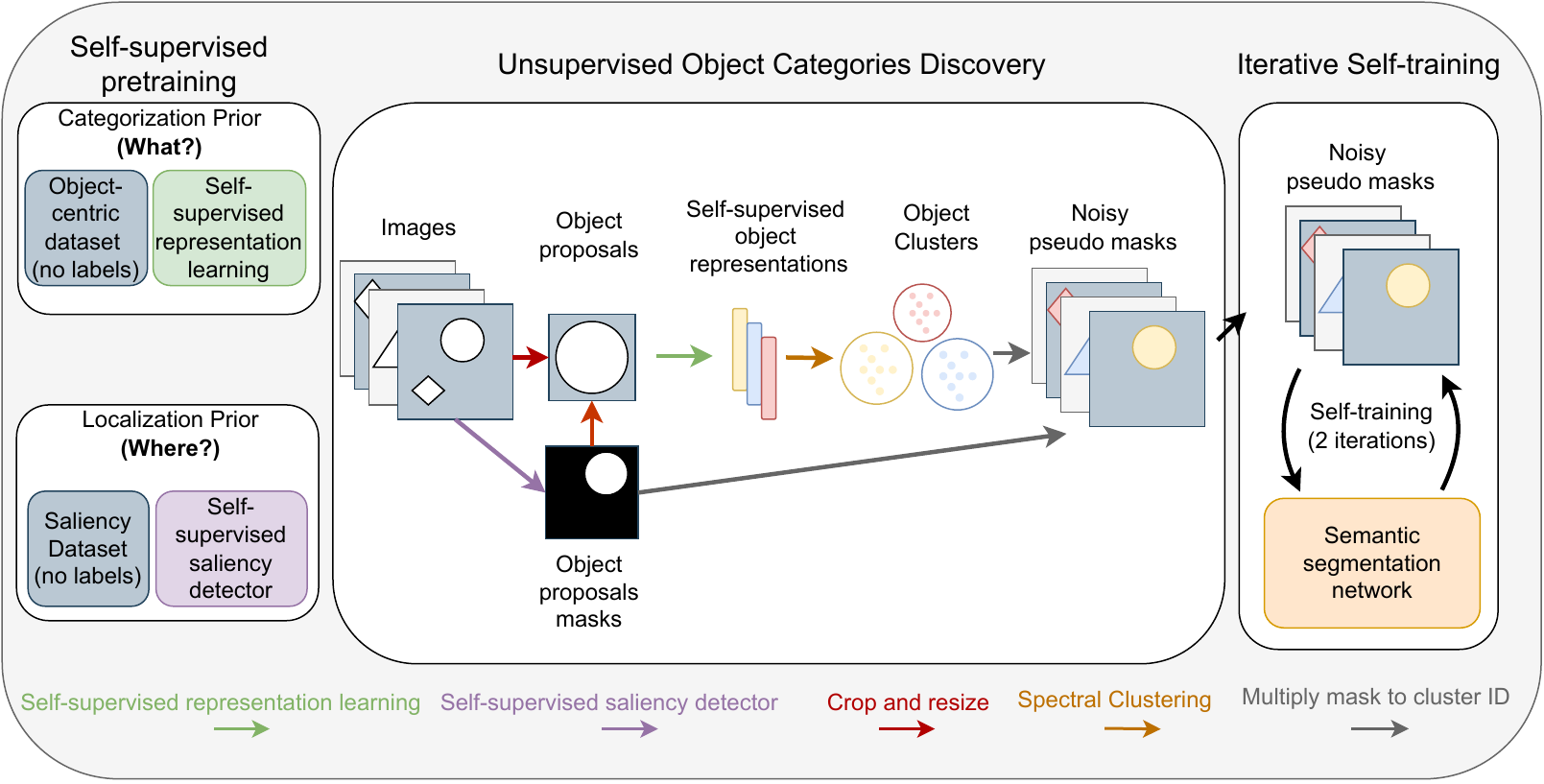}
   \caption{Overview of our self-supervised semantic segmentation framework. First, the self-supervised representation learning network~(e.g.,  DINO~\citep{caronEmergingPropertiesSelfSupervised2021}) and the unsupervised saliency detector~(e.g., DeepUSPS~\citep{Nguyen_Neurips2019}) are trained without manual annotation on object-centric and saliency datasets~(e.g., ImageNet~\citep{imagenet_cvpr09} and MSRA~\citep{mrsa}).
   Next, we use the saliency detector to estimate object proposal masks from the original semantic segmentation dataset. After this, the original images are cropped to the boundaries of object proposal masks and resized. We compute feature vectors within these regions and cluster them with spectral clustering to discover different object categories. We filter the clusters by removing the most uncertain examples. The cluster IDs are combined with the saliency masks to form unsupervised pseudo-masks for self-training of a semantic segmentation network (e.g.,~DeepLabv3). }
   \label{fig:overview}
\end{figure}

\noindent 
Our contributions can be summarized as follows:
\begin{enumerate}[leftmargin=2em,topsep=0em,itemsep=-0.1em]
   \item We propose a strong and simple baseline method (summarized in \Fig{fig:overview}) for unsupervised discovery of object categories and unsupervised semantic segmentation in real-world multi-object image datasets.
   \item We show that unsupervised segmentation can reach quality levels comparable to supervised segmentation 10 years ago~\citep{pascal-voc-2012}.
   This demonstrates that unsupervised segmentation is not only an ill-defined academic playground.
   \item We perform extensive ablation studies to analyze the importance of the individual components in our proposed pipeline, as well as bottlenecks to identify good directions to further improve the quality of unsupervised object discovery and unsupervised semantic segmentation. 
\end{enumerate}

\section{Related Work}
There are several research directions that try to tackle the challenging task of detecting and segmenting objects without any, 
or with only few, human annotations. 

\paragraph{Unsupervised Semantic Segmentation} The first line of work~\citep{vangansbeke2020unsupervised, MaskedAutoencoders2021, choPiCIEUnsupervisedSemantic2021, jiInvariantInformationClustering2019, Hwang2019SegSortSB, oualiAutoregressiveUnsupervisedImage2020a, Hamilton2022Stego, Ke2022UnsupervisedHS} aims to learn dense representations for each pixel in the image and then cluster them (or their aggregation from pixels in the foreground segments) to get each pixel label. While learning semantically meaningful dense representations is an important task itself, clustering them directly to obtain semantic labels seems to be a very challenging task~\citep{jiInvariantInformationClustering2019, oualiAutoregressiveUnsupervisedImage2020a}. 
Thus, usage of additional priors or inductive biases could simplify dense representation learning. PiCIE~\citep{choPiCIEUnsupervisedSemantic2021} incorporates geometric consistency as an inductive bias to facilitate object category discovery. Recently, STEGO~\citep{Hamilton2022Stego} showed that DINO feature correspondences could be distilled to obtain even stronger bias for category discovery. MaskContrast~\citep{vangansbeke2020unsupervised} uses a more explicit mid-level prior provided by an unsupervised saliency detector to learn dense pixel representations. To obtain semantic labels in an unsupervised way, such representations are averaged over saliency masks and clustered. We show that better representations for each mask could be extracted by using off-the-shelf self-supervised representations from DINO~\citep{caronEmergingPropertiesSelfSupervised2021} encoder. 
Recently, DeepSpectral~\citep{melaskyriazi2022deep} proposed to use spectral decomposition of dense DINO features. They suggested over-cluster each image into segments and afterward extracting and clustering DINO representations of such segments while using heuristics to determine the background segment. Those segments represent object parts that could be combined with over-clustering and community detection to improve the quality of pseudo-masks~\citep{Ziegler2022SelfSupervisedLO}.  In contrast, we show that starting from object-centric saliency priors discovered on a simpler dataset provides large benefits for discovering object categories~(see \app{sec:different_sal}). In contrast to estimating pseudo-masks on the full dataset, using only good quality reliable object proposals for each category combined with iterative self-training on expanding datasets additionally decrease biases over the initial pseudo-masks.

\paragraph{Unsupervised Object Discovery (UOD)} UOD is another research direction that also aims to discover object information such as bounding boxes or object masks from images without any human annotations. Recent works on UOD~\citep{Henaff2022ObjectDA, melaskyriazi2022deep, Wang2022TokenCut, Simeoni2021LOST, voLargeScaleUnsupervisedObject2021, zhangObjectDiscoverySingle2020, Vo2020OSD} showed the potential benefit of using the embeddings of pretrained networks (supervised or self-supervised) for both object localization (to the level of the object's bounding box) and object clustering. First, rOSD~\citep{voLargeScaleUnsupervisedObject2021, Vo2020OSD} showed that  supervised features could be used to localize single objects in the image. Next, LOST~\citep{Simeoni2021LOST} proposed a heuristic that relies on self-supervised features 
to localize the most salient object in the image. In contrast to those methods we consider the challenging task of object segmentation, not only object detection. Finally, ~\citet{melaskyriazi2022deep, Wang2022TokenCut} propose to do spectral decomposition of dense DINO features and use of sign of Fiedler eigenvector as criteria for object localization mask. 

\paragraph{Unsupervised Object-centric Representation Learning} Object centric learning assumes that scenes are composed of different objects and aims to learn sets of feature vectors, where each of them binding to one object. 
Unsupervised methods based on single images~\citep{burgess2019monet,greff2019multi,Engelcke2020GENESIS,locatello2020objectcentric, Singh2022SLATE} suffer from single-view ambiguities, which one tries to overcome by exploiting the information in multiple views of a static scene~\citep{chen2021roots}, in a single view of a dynamic scene (i.e., a video) \citep{hsieh2018,kipf2021conditional, Singh2022STEVE} or multiple views of a dynamic scene \citep{Nanbo_Neurips2021}. In contrast
to previous methods and similar to DINOSAUR method~\citep{seitzer2023bridging}, our method exploits unlabeled object-centric datasets to extract object masks and representations.

\section{Self-supervised Semantic Segmentation}

\subsection{Initial Discovery of Object Categories}
\label{subsec:discovery}
Unsupervised decomposition of complex, multi-object scenes into regions that correspond to the present objects categories is hard and largely ill-defined as it is possible to decompose scene with different levels of granularity obtaining several valid decompositions for the same scene (e.g., a person could be considered as one object or additionally decomposed to body parts). However, it is unnecessary to correctly decompose all images of a dataset to kickstart unsupervised object discovery. In this paper, we show that it is sufficient to exploit simple images in a dataset in order to discover \textit{some} objects and their categories. This works particularly well due to intrinsic properties of natural images and photos made by humans. One such property is that the most salient region of the image often corresponds to a single distinct object.

\begin{algorithm}[tb]
\caption{Object Categories  Discovery for Unsupervised Pseudo-Masks Estimation}\label{alg:main1}%
\resizebox{0.9\textwidth}{!}{%
\begin{minipage}{1.0\textwidth}
  \begin{algorithmic}
  \State \textbf{Given}: $N$ images $x_i$, self-supervised salient regions' segmentation network $L$ with binary threshold $\theta$, self-supervised representation learning method $C$, percentage of proposals to filter $p$.
  
  \vspace{2mm}
  \State \textbf{Step 1}: Obtain binary object proposal masks $m_i$ by  $s_i = L(x_i) > \theta $ and object proposal regions $o_i = \crop(x_i, s_i)$.
  \State \textbf{Step 2}: Compute object representations $r_i$ of object proposal regions $o_i$, $r_i = C(o_i)$
  \State \textbf{Step 3}: Cluster object proposal representations $r_i$ using spectral clustering to assign cluster ID $t_i$ for each object proposal $o_i$.
   \State \textbf{Step 4}: Filter $p$ percents of the most uncertain object proposals for each discovered cluster (proposals  with the largest distance to the cluster center in the eigenvalue embedding).
  \State \textbf{Step 5}: Combine cluster IDs $t_i$ with object proposal masks $s_i$ to obtain initial pseudo-masks $m_i$. 
  
  \vspace{2mm}
  \State \textbf{Return:} Noisy object segmentation pseudo-masks $m_i$.
  \end{algorithmic}
  \end{minipage}%
}%
\end{algorithm}
\paragraph{Self-supervised Object Localization}
Similar to MaskContrast~\citep{vangansbeke2020unsupervised}, we propose to retrieve a set of object mask proposals for the images in our dataset by using an unsupervised saliency estimator. In particular, we were using the DeepUSPS~\citep{Nguyen_Neurips2019} model as an unsupervised saliency estimator. DeepUSPS was trained on MSRA~\citep{mrsa} in an unsupervised way exploiting the bias towards simple scenes with often homogeneously textured background of the MSRA dataset, as well as the hard-coded saliency priors of classical (non-learning-based) saliency methods. To further improve the estimator's transfer ability to more complex datasets like PASCAL VOC and MS COCO, we trained another saliency model, BasNet~\citep{Qin_2019_CVPR}, on the saliency masks generated by DeepUSPS. Some examples of saliency detector masks are presented in \Sec{sec:ablations}; see \Fig{fig:vis_masks}. In addition, we studied performance of our method with original  DeepUSPS masks and with DeepSpectral saliency masks in \app{sec:sal}.

\paragraph{Self-supervised Representation Learning}
The self-supervised feature learning technique DINO~\citep{caronEmergingPropertiesSelfSupervised2021} exploits the dataset bias of ImageNet, which mostly shows a single object in the center of the image. DINO uses, among other transformations, the multi-crop strategy to link local patterns of the same object instance to its global pattern in the embedding. This leads to a feature representation that tends to have a similar embedding for patterns from the same category.

We start extraction of the object representation by cropping the image to the borders of the saliency mask and resizing the obtained crop to $256\times256$ resolution.
Next, we feed the object proposal into the Vision Transformer~(ViT)~\citep{dosovitskiy2020image} architecture pretrained in a self-supervised way with DINO. The feature vector of the CLS token from the last layer is used as object representation. As CLS token attention values from  the last layer of DINO were shown to attend to foreground objects~\citep{caronEmergingPropertiesSelfSupervised2021}, the obtained CLS token features are implicitly aggregating object related information from the object proposal.

\paragraph{Discovery of Semantic Categories}
We cluster the feature vectors obtained per image with spectral clustering~\citep{Luxburg2007ATO}. Thanks to the saliency masks, most of the feature vectors are based on foreground patterns and disregard the background, i.e., they become object-centric. Even though this is clearly not the case for all images, either because there are salient regions in the background or because the image shows multiple objects from different categories, there are enough good cases for spectral clustering to yield clusters that are dominated by a single object category; see \Fig{fig:vis_clustering}. As we show in \Tab{tab:ablations}, this clustering of features within the saliency masks already yields unsupervised object discovery results beyond the state of the art.

\paragraph{Filtering of Cluster Samples}
Since neither the salient regions nor DINO features are perfect, we must expect several outliers within the clusters. We tested the simple procedure to filter the most uncertain samples of each cluster and discard them. We measure uncertainty by the distance to the cluster's mean in the spectral embedding of the Laplacian eigenmap~\citep{Luxburg2007ATO}. In \Fig{fig:vis_clustering} we show that such examples are often failure cases of the saliency detector, such as parts of background that are not related to any category. In addition, we study sensitivity of COMUS algorithm in \app{sec:clustering_parameters}, showing that \method performs comparably well when the percentage of filtered examples varies from $20\%$ to $40\%$. We refer the reader to the \Alg{alg:main1} for the detailed pseudocode of object categories discovery part of the \method method.

\subsection{Unsupervised Iterative Self-training with Noisy Pseudo-Masks}
\begin{algorithm}[tb]
\caption{Self-training with Noisy Pseudo-Masks}
\label{alg:self_training}
\resizebox{0.9\textwidth}{!}{%
\begin{minipage}{1.0\textwidth}
  \begin{algorithmic}
  \State \textbf{Given}: $N$ images $x_i$ with clustering pseudo-masks $m_i$, external $M$ images $x_j$ for self-training
  
  \vspace{2mm}
  \State \textbf{Step 1}: Train a Teacher network $\theta_{t}$ (with prediction function $f$) on images with unsupervised pseudo-masks by minimizing the total loss $\mathcal{L}$ for object segmentation:
  $$\theta_{t}^{*} = \argmin_{\theta_t} \frac{1}{N}\sum_{j=1}^{N}\mathcal{L}(m_j, f(x_j, \theta_{t})).$$
  \State \textbf{Step 2}: Generate new pseudo-masks $\widetilde{m}_j$ for all unlabeled images $x_j$ (e.g., images from PASCAL VOC \emph{trainaug} set).
  \State \textbf{Step 3}: Train a Student network $\theta_{s}$ on images and new pseudo-masks ($x_j$, $\widetilde{m}_j$):
$$\theta_{s}^{*} = \argmin_{\theta_s} \frac{1}{N+M}\sum_{j=1}^{N+M}\mathcal{L}(\widetilde{m}_j, f(x_j, \theta_{s})).$$

\vspace{2mm}
   \State \textbf{Return:} Semantic segmentation network $\theta_{s}^*$.
  \end{algorithmic}
\end{minipage}
}
\end{algorithm}

As discussed above, the clustering of feature vectors extracted from within saliency masks makes several assumptions that are only satisfied in some of the samples of a dataset. While this is good enough to get the object discovery process kickstarted, it is important to alleviate these assumptions in order to extend to more samples. In particular, we implicitly relied on the image to show only objects from one category (otherwise the feature vector inside the saliency mask comprises patterns from different categories) and on the correct localization of the object boundaries.

To extend also to multi-object images and to improve the localization of object boundaries, we propose using the masks with the assigned cluster IDs as initial pseudo-labels for iterative self-training of a semantic segmentation network. Self-training is originally a semi-supervised learning approach that uses labels to train a teacher model and then trains a student model based on the pseudo-labels generated by the teacher on unlabeled data~\citep{Xie2020SelfTrainingWN}. Similar self-training methods were shown to be effective also in semantic segmentation~\citep{Chen2020LeveragingSL, Zhu2020ImprovingSS}. In this paper, we use the unsupervised pseudo-masks from \sec{subsec:discovery} to train the teacher.
In our experiments, we used the network architecture of DeepLabv3~\citep{Chen2017RethinkingAC}, but the method applies to all architectures. Since large architectures like DeepLabv3 are typically initialized with an ImageNet pretrained encoder, we also use a pretrained encoder for initialization. However, since we want to stay in the purely unsupervised training regime, we use  self-supervised DINO pretraining.

Once the semantic segmentation network is trained on the pseudo-masks, it can predict its own masks. In contrast to the saliency masks, this prediction is not limited to single-object images. Moreover, the training can consolidate the information of the training masks and, thus, yields more accurate object boundaries. Since the masks of the segmentation network are on average better than the initial pseudo-masks, we use them as pseudo-masks for a second iteration of self-training. In addition, if such masks are obtained from unseen images of an extended dataset, the predictions of the segmentation network are not overfitted to the initial pseudo-masks and thus are an even better supervision signal. We refer to the pseudocode in \Alg{alg:self_training} for an overview of iterative self-training. In addition, \Tab{tab:ablations} in \sec{sec:ablations} shows that both the initial self-training (Step~$1$ in \Alg{alg:self_training}) and the second iteration~(Step~$3$ in \Alg{alg:self_training}) of self-training improve~results. 

\section{Experiments}

\paragraph{Evaluation Setting}
We tested the proposed approach on two semantic object segmentation datasets, PASCAL VOC~\citep{pascal-voc-2012} and  MS COCO~\citep{Lin2014MicrosoftCC}. These benchmarks are classically used for supervised segmentation. In contrast, we used the ground truth segmentation masks only for testing but not for any training. 
We ran two evaluation settings. For the first, we created as many clusters as there are ground truth classes and did one-to-one Hungarian matching~\citep{kuhn1955} between clusters and classes. For the second, we created more clusters than there are ground truth classes and assigned the clusters to classes via majority voting, i.e, for each cluster we chose the class label with most overlap and assigned the cluster to this class. 
In both cases we used IoU as the cost function for matching and as the final evaluation~metric.

Hungarian matching is more strict, as it requires all clusters to match to a ground truth class. Hence, reasonable clusters are often marked as failure with Hungarian matching; for instance, the split of the dominant person class into sitting and standing persons leads to one cluster creating an IoU of 0. This is avoided by majority voting, where clusters are merged to best match the ground truth classes. However, in the limit of more and more clusters, majority voting will trivially lead to a perfect result.
When not noted otherwise, we used Hungarian matching in the following tables. 
We report mean intersection over union (mIoU) as evaluation metric.

\paragraph{Implementation Details}
We used pretrained DINO features with the DINO architecture released in DINO's official GitHub\footnote{\url{https://github.com/facebookresearch/dino}}. In particular, we used DINO with patch size 8 that was trained for 800 epochs on ImageNet-1k without labels.
For the saliency masks, we used the BasNet weights pretrained on predictions from DeepUSPS released by MaskContrast's official GitHub\footnote{\url{https://github.com/wvangansbeke/Unsupervised-Semantic-Segmentation}} (see folder \texttt{saliency}). All parameters of spectral clustering and self-training are described in \app{sec:parameters}.

\subsection{PASCAL VOC Experiments}
\begin{table}[tbp]
\footnotesize
\renewcommand{\arraystretch}{1.2}
\centering
\caption{Comparison to prior art and iterative improvement via  self-training 
(evaluated by IoU after Hungarian matching) 
on the PASCAL 2012 \emph{val} set. The results for SwAV and IIC methods are taken from MaskContrast paper. \method results are mean $\pm$ standard dev.\@ over 5 runs. 
}
\label{tab:iterative}
\begin{adjustbox}{max width=0.48\textwidth}
  \setrow{\bfseries}
  \begin{tabular}{r@{\hskip1em} *{1}{c}@{\hskip1em}c<{\clearrow}}
  \toprule
  \textbf{Method} & \textbf{mIoU}
     \\ \midrule

  Colorization~\citep{zhang2016colorful}& 4.9\\
  IIC~\citep{jiInvariantInformationClustering2019} & 9.8\\
  SwAV~\citep{caron2020unsupervised} & 4.4\\
  MaskContrast~\citep{vangansbeke2020unsupervised} & 35.1\\
  DeepSpectral~\citep{melaskyriazi2022deep} & 37.2 $\pm$ 3.8\\
  DINOSAUR~\citep{seitzer2023bridging} & 37.2 $\pm$ 1.8\\
  Leopart~\citep{Ziegler2022SelfSupervisedLO} & 41.7\\

  \midrule
  \setrow{\bfseries}
  Pseudo-masks (Iteration 0) & 43.8 $\pm$ 0.1\\
  \method (Iteration 1) & 47.6 $\pm$ 0.4\\
  \method (Iteration 2) & \textbf{50.0 $\pm$ 0.4}\\
  \bottomrule
  \end{tabular}
\end{adjustbox}
\end{table}

\begin{table}[tbp]
\renewcommand{\arraystretch}{1.2}
\centering
\caption{\method performance on PASCAL VOC 2007 \textit{test} (evaluated by IoU after Hungarian matching). The test data was never seen during self-learning or validation.}
\label{tab:test}
\begin{adjustbox}{max width=\textwidth}
  \setrow{\bfseries}
  \begin{tabular}{r@{\hskip1em}*{21}{c}@{\hskip1em}c<{\clearrow}}
  \toprule
   & \tableIcons & \textbf{mIoU} \\ \midrule
   DeepSpectral & 77.3& 40.2& 0.0& 78.2& 25.0& 6.0& 65.7& 50.7& 82.7& 0.0& 43.6& 24.5& 54.4& 63.5& 31.5& 20.6& 2.3& 0.0& 9.4& 77.0& 0.1& 35.8 \\ \midrule
  \method  & 84.3& 38.3& 30.9& 51.4& 47.1& 39.9& 66.3& 54.7& 67.7& 0.0& 60.2& 21.3& 54.3& 57.9& 62.7& 45.4& 9.0& 72.2& 13.8& 81.5& 43.4& 47.7 \\
  \bottomrule
  \end{tabular}
\end{adjustbox}
\vspace{-1em}
\end{table}

\label{sec:pascal_exp}
PASCAL VOC 2012~\citep{pascal-voc-2012} comprises $21$ classes -- $20$ foreground objects and the background. 
First, to qualitatively validate that the obtained clusters correspond to true object categories, we visualize the t-SNE embedding~\citep{JMLR:v9:vandermaaten08a} of DINO representations showing that clusters correspond to different object categories (see %
\fig{fig:vis_clustering}).
Further, we quantitatively confirmed that saliency masks with assigned cluster ID (pseudo-masks) produce state-of-the-art unsupervised semantic segmentation on PASCAL VOC and outperforms the MaskContrast method that learns dense self-supervised representations; see Table \ref{tab:iterative} (Iteration~0~row). 

For self-training the DeepLabv3 architecture, we initialized the encoder with a ResNet50 pretrained with DINO (self-supervised) on ImageNet and finetuned the whole architecture on the pseudo-masks we computed on the PASCAL 2012 \textit{train} set. This increased the performance from $43.8$\% mIoU to $47.6$\% mIoU, see \Tab{tab:iterative} (Iteration 1), which supports our consideration of bootstrapping from the original pseudo-masks. In particular, it allows us to segment objects in multi-category images. 

\begin{table}[tbp]
\caption{Unsupervised semantic segmentation before and after self-learning evaluated by mIoU after Hungarian matching on the MS COCO \emph{val} set. As discovered object category we count those categories with an IoU $>20\%$ from all 81 categories. Also, we show IoU for categories that have corresponding cluster (i.e., with IoU larger than zero).}
\label{tab:unsup_coco}
\centering
\hspace{0.24cm}
\begin{adjustbox}{max width=0.8\textwidth}
\begin{tabular}{l ccccccc}
\toprule
& all  & {} &   \multicolumn{2}{c}{discovered (with IoU$\ge 20\%$)} & {} & \multicolumn{2}{c}{have cluster (with IoU$>0\%$)} \\
\cmidrule{2-2} \cmidrule{4-5} \cmidrule{7-8} 
 & mIoU  && number &  mIoU && number &  mIoU \\
\midrule
Pseudo-masks &	18.2 && 33 & 36.6 && 73 & 20.2   \\
\method \hspace{2mm} &  19.6 && 34	& 40.7  && 60 & 26.5 \\
\bottomrule
\end{tabular}
\end{adjustbox}
\end{table}

\begin{table}[tbp]
\renewcommand{\arraystretch}{1.2}
\centering
\caption{Transfer from PASCAL VOC to MS COCO for the 21 PASCAL VOC classes. 
Training on the simpler PASCAL dataset yields better performance on COCO than learning on COCO itself while both \method runs perform better than DeepSpectral.}
\label{tab:trasnfer}
\begin{adjustbox}{max width=\textwidth}
  \setrow{\bfseries}
  \begin{tabular}{r@{\hskip1em}*{21}{c}@{\hskip1em}c<{\clearrow}}
  \toprule
   & \tableIcons & \textbf{mIoU}\\ \midrule	
   DeepSpectral & 71.6& 42.4& 0.0& 51.6& 10.1& 0.7& 54.5& 22.9& 66.9& 1.4& 2.3& 20.1& 35.7& 48.3& 39.2& 16.3& 0.0& 29.4& 1.9& 40.2& 7.0& 26.8 \\
   \midrule	
  \method~(trained on PASCAL)  & 79.5 & 40.7& 12.4& 31.9& 25.7& 14.0& 50.6& 12.1& 56.1& 0.0& 31.0& 20.1& 47.6& 39.6& 40.6& 43.5& 6.8& 47.6& 8.0& 39.7& 22.8& 31.9 \\
  \method~(trained on COCO) & 76.5 & 39.9& 28.2& 29.7& 34.3& 0.1& 56.8& 6.8& 34.9& 0.7& 50.2& 4.4& 42.1& 38.7& 48.4& 15.1& 0.0& 54.4& 0.0& 40.4& 2.6 & 28.8 \\
  \bottomrule
  \end{tabular}
\end{adjustbox}
\vspace{-1em}
\end{table}

Successively, we added one more iteration of self-learning on top of the pseudo-masks on the PASCAL 2012 \textit{trainaug} set. The PASCAL 2012 \textit{trainaug} set ($10 582$ images) is an extension of the original \textit{train} set ($1 464$ images) \citep{pascal-voc-2012, Hariharan2011SemanticCF}. It was used by previous work on fully-supervised~\citep{Chen2018DeepLabSI} and unsupervised~\citep{vangansbeke2020unsupervised} learning. The second iteration of self-training further improves the quality to $50.0\%$ mIoU; see \Tab{tab:iterative} (Iteration 2). In particular, it allows us to make multi-category predictions on images from the validation set unseen during self-supervised training (\fig{fig:vis}). Accordingly, the method also yields good results on the PASCAL VOC 2007 official test set; see \Tab{tab:test}.

\subsection{MS COCO Experiments}
\label{sec:coco_exp}
We further evaluated our method on the more challenging COCO dataset~\citep{Lin2014MicrosoftCC}. It  focuses on object categories that appear in context to each other and has $80$ things categories. We transform the instance segmentation masks to category masks by merging all the masks with the same category together. Our method is able to discover $34$ categories with more than $20\%$ IoU. Among those categories, we obtained an average IoU of $40.7\%$; see \Tab{tab:unsup_coco}.

Additionally, we studied the transfer properties of \method under a distribution shift. To this end, we self-trained our \method model on the PASCAL VOC dataset and then tested this model on the same $20$ classes on the MS COCO dataset. 
The results in \Tab{tab:trasnfer} show that the transfer situation between datasets is quite different from supervised learning: training on the PASCAL VOC dataset and testing on COCO yields better results than training in-domain on COCO (see \fig{fig:transfer} in Appendix). This is because the PASCAL VOC dataset contains more single object images than MS COCO, which makes self-supervised learning on PASCAL VOC easier. 
This indicates that the order in which data is presented plays a role for unsupervised segmentation. 
Starting with datasets that have more single-object bias is advantageous over starting right away with a more complex dataset. 

\begin{figure}
\begin{minipage}[t]{0.60\textwidth}
    \centering
    \captionof{table}{Ablation experiment to identify the effect of individual components of the unsupservised learning process. 
    }
\label{tab:ablations}
\begin{tabular}{P{17mm}|P{13mm}|P{13mm}|P{13mm}|P{10mm}}
\textbf{Laplacian Eigenmap} & \textbf{Self-training} & \textbf{Filtering} &\textbf{2nd self-training}  & \textbf{mIoU} \\
\toprule
\xmark & \xmark & \xmark & \xmark & 42.8 \\
\checkmark & \xmark & \xmark & \xmark & 43.8 \\
\checkmark & \checkmark & \xmark & \xmark & 47.1 \\
\checkmark & \checkmark & \checkmark & \xmark & 47.6 \\
\checkmark & \checkmark & \checkmark & \checkmark & 50.0 \\
\bottomrule
\end{tabular}
    \centering
        \captionof{figure}{Visualization of foreground masks obtained with different foreground segmentation methods. 
        }
    \includegraphics[width=\linewidth]{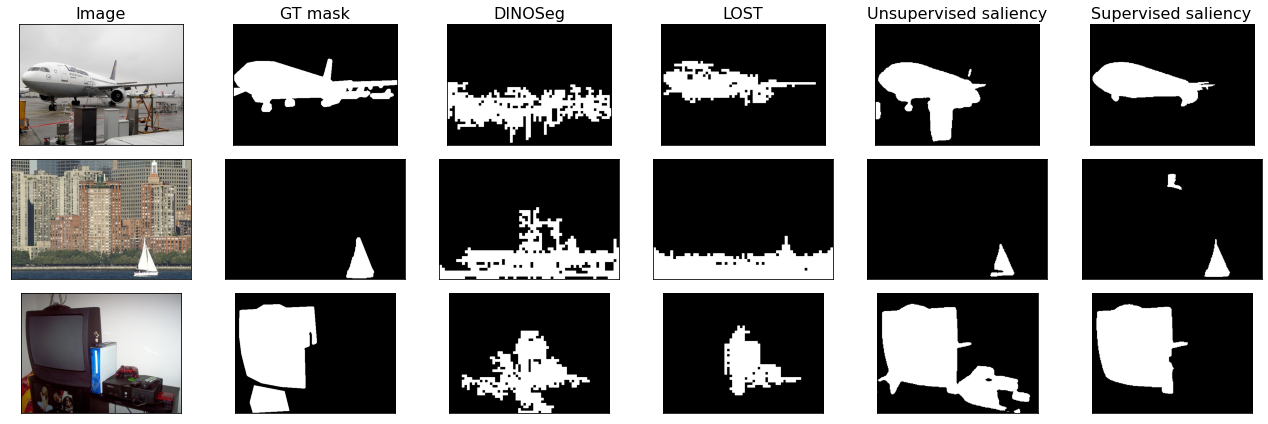} \\
    \label{fig:vis_masks}
\end{minipage}\hfill
\begin{minipage}[t]{0.37\textwidth}
\centering
\captionof{table}{Comparison of COMUS performance with different feature extractors on PASCAL VOC.}
\label{tab:feature_extractors}
\setrow{\bfseries}
\begin{tabular}{l@{\hskip1em}*{1}{c}@{\hskip1em}c<{\clearrow}}
\toprule
 & \textbf{mIoU}
 \\ \midrule
\method with SwAV & 28.6\\
\method with iBOT & 43.8\\
\midrule
\setrow{\bfseries}
\method  with DINO & \textbf{50.0}\\
\bottomrule
\end{tabular}

\vspace{1.5em}
\centering
\captionof{table}{Comparison between different class-agnostic foreground segmentation methods. 
}
  \begin{tabular}{l@{\hskip1em}*{1}{c}@{\hskip1em}c<{\clearrow}}
  \toprule
   & \textbf{IoU} \\ \midrule
  Unsupervised saliency  & \textbf{51.0} \\
   LOST 
   & 34.8 \\
  DINOSeg 
   & 24.5 \\
  \midrule
  Supervised saliency & 60.5 \\
  \bottomrule
  \end{tabular}
\label{tab:mask_quality}
\end{minipage}
\vspace{-1em}
\end{figure}

\subsection{Analysis}
\label{sec:ablations}
\paragraph{Ablation Study}
To isolate the impact of single components of our architecture, we conducted various ablation studies on PASCAL VOC; see \Tab{tab:ablations}. All proposed components have a positive effect on the result: spectral clustering that additionally computes Laplacian eigenmap before k-means clustering yields better results than k-means clustering~(see \app{sec:clustering_parameters} for detailed analysis on clustering method choice and sensitivity of its parameters); self-training is obviously important to extend to multi-category images; filtering the most distant samples from a cluster followed by the second iteration of self-training on the much larger \emph{trainaug} set gives another strong boost. 
\paragraph{Choice of Categorization Prior}
Next, we investigated how \method works with different self-supervised representation learning methods. \method performs best with ViT based features extractors such as DINO~\citep{caronEmergingPropertiesSelfSupervised2021} and iBOT~\citep{zhou2021ibot}, while its performance is significantly worse for SwAV method~\citep{caron2020unsupervised}  based on ResNet architecture. We further show that clustering ability of categorization method could be evaluated on ImageNet images clustering where self-supervised methods were originally trained on (see \app{sec:features_quality}).

\paragraph{Quality and Impact of Saliency Masks}
In \Tab{tab:mask_quality}, we compare the quality of the used unsupervised saliency mask detector with other recently proposed detection methods. We report the IoU for the foreground class while using different methods for foreground object segmentation. In particular, we evaluated segmentation masks proposed in LOST~\citep{Simeoni2021LOST} that uses DINO keys correlation between features, and DINOSeg~\citep{Simeoni2021LOST, caronEmergingPropertiesSelfSupervised2021}, which uses the attention weights from the last layer of the DINO network, to construct foreground object segmentation masks (see \Fig{fig:vis_masks} for examples of predictions by foreground segmentation methods that we consider). The chosen unsupervised saliency based on DeepUSPS and BASNet outperforms both LOST and DINOSeg with a large margin showing the importance of additional localization prior in contrast with relying only on DINO as both categorization and localization prior. In addition, we show how the quality of saliency masks proposal affects \method performance in \app{sec:sal}.

\paragraph{Limitations and Future Work Directions}
Although \method shows very promising results on the hard tasks of unsupervised object segmentation, there are a number of limitations, as to be expected. 
First, although we reduced the dependency on the quality of the saliency detector via successive self-training, the approach still fails to segment objects that are rarely marked as salient~(see \Fig{fig:vis_falures} in \app{sec:falure_modes}). 
Second, while the bootstrapping via self-learning can correct some mistakes of the initial discovery stage, it cannot correct all of them and can be itself biased towards self-training data~(see~\app{sec:n_objects}). 
Third, we fixed the number of clusters in spectral clustering based on the known number of categories of the dataset. While \method works reasonably with larger number of clusters~(we refer to \app{sec:over_clustering} for over-clustering experiments), in a fully open data exploration scheme, the optimal number of clusters should be determined automatically. 

\section{Conclusion}
\looseness=-1In this work, we presented a procedure for semantic object segmentation without using any human annotations clearly improving over previous work.
As any unsupervised segmentation method requires some biases to be assumed or learned from data, we propose to use object-centric datasets on which localization and categorization priors could be learned in a self-supervised way. We show that combining those priors together with an iterative self-training procedure leads to significant improvements over previous approaches that rely on dense self-supervised representation learning. This combination reveals the hidden potential of object-centric datasets and allows creating a strong baseline for unsupervised 
segmentation methods by leveraging and combining 
learned priors.

While research on this task is still in its infancy, our procedure allowed us to tackle a significantly more complex dataset like MS COCO for the first time. Notably, on PASCAL VOC we obtained results that match the best supervised learning results from 2012, before the deep learning era. Hence, the last ten years of research not only have yielded much higher accuracy based on supervised learning, but also allow us to remove all annotation from the learning process.

\newpage

\section*{Acknowledgments}
We would like to thank Maximilian Seitzer and Yash Sharma for insightful discussions and practical advice. In addition, we thank Vit Musil for PASCAL VOC class icons.

\section*{Reproducibility Statement}
\Alg{alg:main1} and \Alg{alg:self_training} contain pseudocode for both parts of the \method method.
\App{sec:parameters} contains detailed information about the \method architecture and all hyperparameters used for \method training.
\App{sec:datasets} contains information about the datasets used in the paper.
The COMUS implementation code is located here: \href{https://github.com/zadaianchuk/comus}{https://github.com/zadaianchuk/comus}. 

\bibliography{main}
\bibliographystyle{iclr2023_conference}

\appendix
\newpage
\begin{center}\LARGE\textsc{Appendix}\end{center}

\section{Over-clustering}
\label{sec:over_clustering}
\begin{figure}[ht]
   \includegraphics[width=\linewidth]{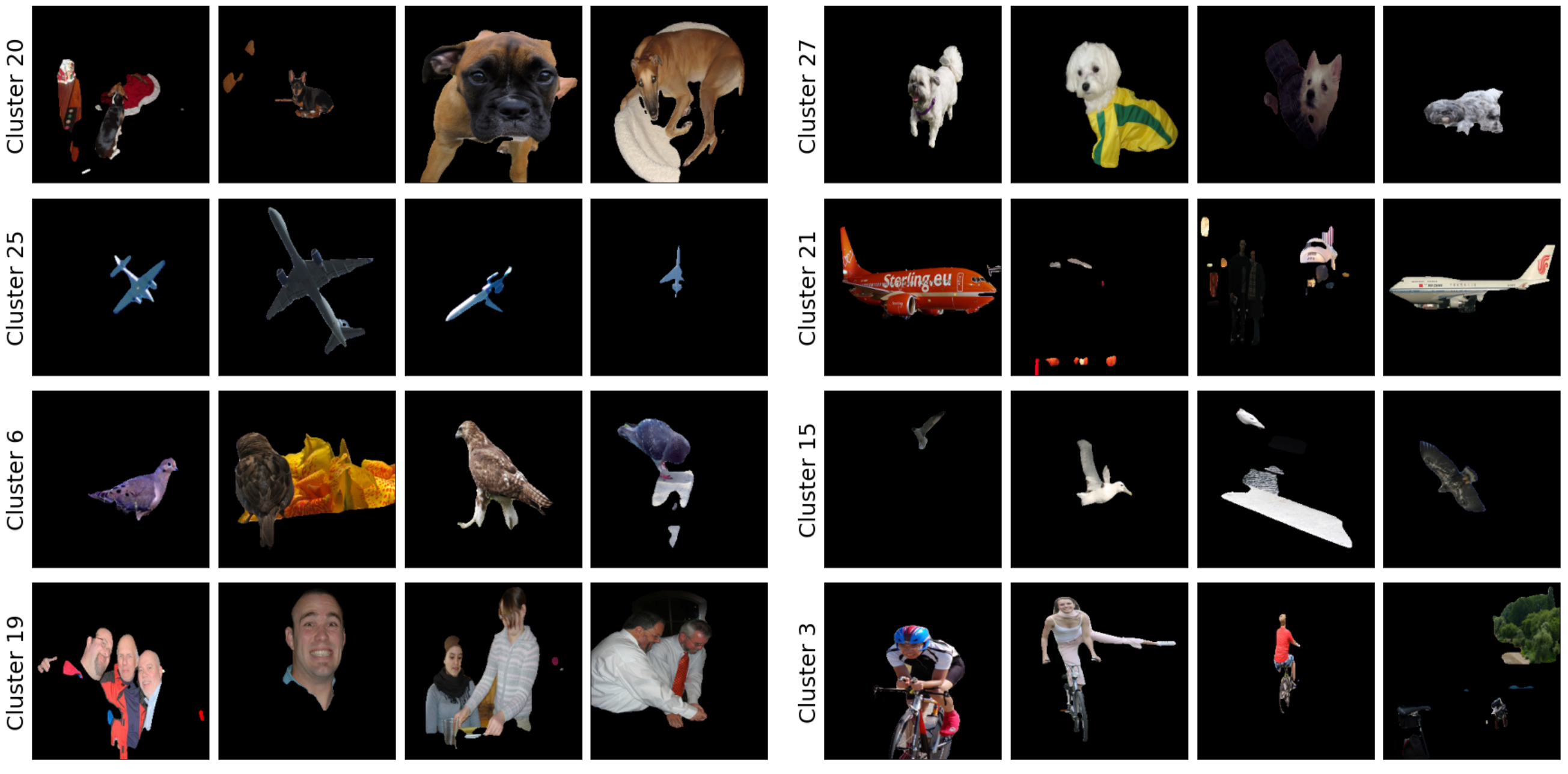}
   \caption{Visualization of discovered subcategories on PASCAL VOC \textit{val} set after clustering of self-supervised representations into 30 clusters. The pseudo-masks were randomly sampled for each cluster. Each row shows two clusters of the same category. The clusters have clear semantic interpretation, such as different dog breeds, flying or staying on land airplanes. }
   \label{fig:vis_subcategories}
\end{figure}

\begin{table}[h]
\caption{Over-clustering results on PASCAL VOC evaluated with mIoU after majority voting. We present the results for 30 clusters, whereas also include the results for 50 clusters for comparison with MaskContrast~\citep{vangansbeke2020unsupervised}.}
\label{tab:overclustering_pascal_voc}
\centering
\hspace{0.24cm}
\begin{tabular}{l c c}
\toprule
 & 30 clusters & 50 clusters \\
\midrule
MaskContrast & -  & 41.4 \\ \midrule
\method~(Iteration 1) & 	49.3 & 	46.9 \\
\method~(Iteration 2)	& 52.6 & 51.0 \\
\bottomrule
\end{tabular}
\end{table}
As the number of discovered clusters could be different from the number of human-defined categories, we ran our process with a larger number of clusters than there are ground truth categories (over-clustering). Each cluster was matched to the ground truth category with the highest IoU, i.e., each category can have multiple clusters being assigned to it. This kind of matching avoids penalization of reasonable subcategories (see \fig{fig:vis_subcategories}). Discovery of subcategories is a strong motivation for using unsupervised methods.  \Tab{tab:overclustering_pascal_voc} shows that also under this evaluation protocol we obtain better results. For comparison to MaskContrast, we also included results with 50 clusters, showing that \method outperforms MaskContrast in the 50 clusters setting. 

Self-training from pseudo-masks with a larger number of categories decreased the performance slightly. Potentially the segmentation network has difficulties learning patterns when several clusters have the same semantic interpretation. In case of fixed saliency masks, this evaluation setting yields better numbers and becomes trivial as the number of clusters becomes very large. 

\newpage 
\section{Sensitivity of the COMUS parameters}
\label{sec:clustering_parameters}
\begin{figure}[t]
  \centering
   \caption{(a) COMUS (Iteration 1) performance with different clustering methods. (b) Effect of "\% filtered" on final performance of COMUS. The results are mean $\pm$ standard dev. over 5 runs.}
  \includegraphics[width=0.7\linewidth]{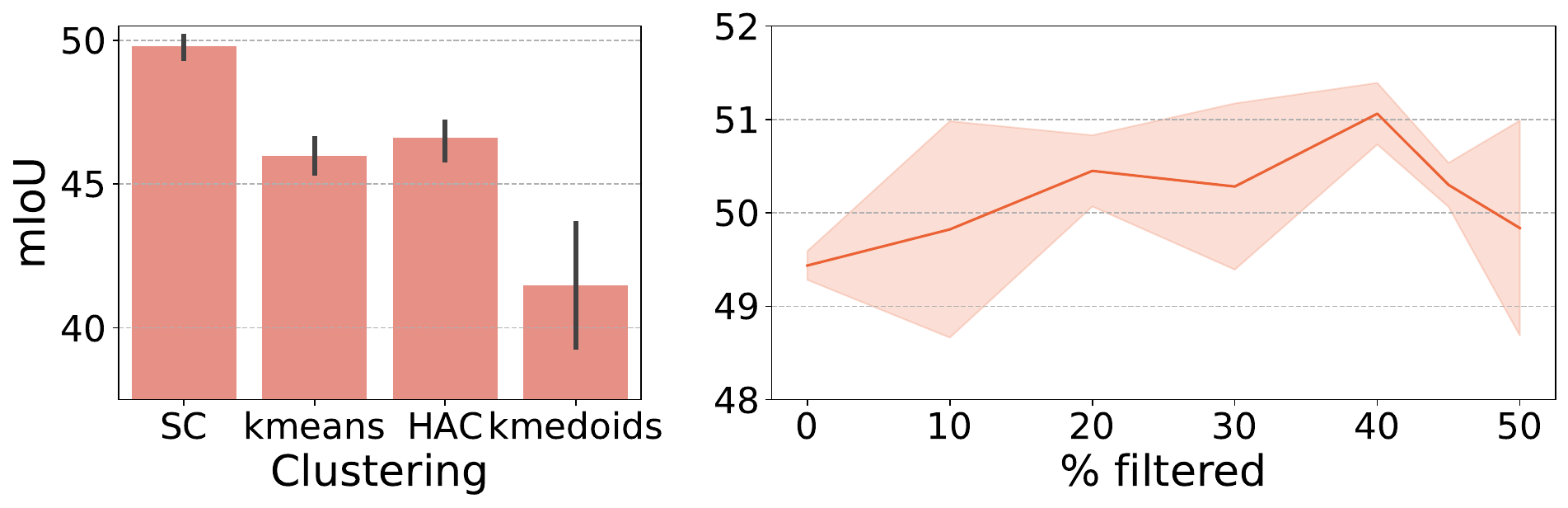}
   \label{fig:filtered}
\end{figure}

First, we compare \method performance with different clustering methods. In \hyperref[{fig:filtered}]{Figure~\ref*{fig:filtered}a} we show that SC provides the best supervision signal for \method. Additionally, we investigate the sensitivity of COMUS to the choice of $\%$ of filtered examples (see \hyperref[{fig:filtered}]{Fig.~\ref*{fig:filtered}b}), showing that [20\%-40\%] of filtered examples leads to comparably good performance. We observe that larger $\%$ of filtered examples leads to drop in performance, potentially due to small size of the obtained dataset for training the segmentation network. 

Next, we look at the sensitivity of \method to the choice of SC parameters. In \Tab{tab:clustering_params} we show that SC is not sensitive to the choice of the number of neighbors and \texttt{n\_init} parameters. SC is sensitive to the number of eigenvectors, however, the default value (equal to the number of clusters) shows very good performance. 
Similar to other works~(e.g.~\cite{vangansbeke2020unsupervised, melaskyriazi2022deep}), we used the same number of clusters as annotated categories~(needed for quantitative evaluation). 

\begin{figure}[h]
\begin{minipage}[t]{0.4\textwidth}
  \centering
  
   \captionof{figure}{Number of self-training iterations in \method training. The results are mean $\pm$ standard dev. over 5 runs.}
  \vspace{-1em}
  \includegraphics[width=\linewidth]{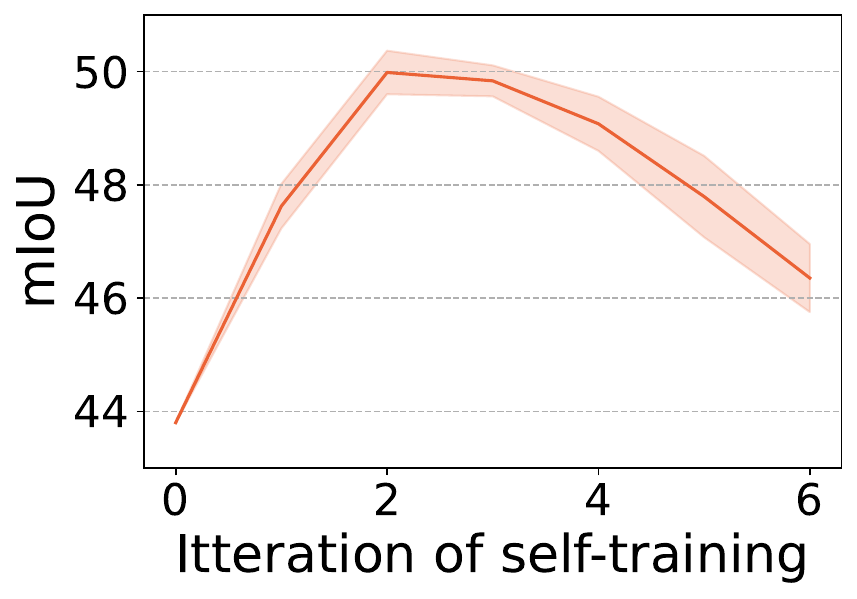}
   \label{fig:itterations}
\end{minipage}
\hfill
\begin{minipage}[t]{0.58\textwidth}
\centering
\captionof{table}{Spectral Clustering parameters study, performance after the first iteration.}
\label{tab:clustering_params}
\centering
\vspace{0.5em}
\begin{tabular}{cccc}
\toprule
              & \texttt{n\_neighbors} & \texttt{n\_components} & \texttt{n\_init}     \\ 
\midrule
Range         & $[20-50]$                        & $ [10-40] $                     & $[10-60]$                           \\ 
\midrule
mIoU & $ 46.0 \pm 1.3$                  & $ 38.8 \pm 6.2 $               & $47.6 \pm 1.0$ \\ 
\bottomrule
\end{tabular}

\end{minipage}

\end{figure}

Finally, we also study the effect of additional iterations of self-training iterations on \texttt{trainaug} part of PASCAL VOC dataset. We find that trained for two or three iterations \method performs similarly, while for even more self-training iterations we observe slow decrease of the performance (\fig{fig:itterations}). While we were performing additional self-training iterations on the same dataset, in the future work, it is interesting to study how our method performs in the open-ended regime where new data is available for each iteration.

\section{Self-supervised features quality}
\label{sec:features_quality}
As was shown in the original paper~\citep{caronEmergingPropertiesSelfSupervised2021}, DINO features are demonstrating excellent performance for k-NN classification (78.3\% top-1 ImageNet accuracy), which reveals the quality of the feature space for clustering. In contrast, other self-supervised methods require fine-tuning of the last layer~\citep{caron2020unsupervised, He2020MomentumCF} or several last layers~\citep{MaskedAutoencoders2021}. 
We further confirmed (see \Tab{tab:features}) that DINO performs significantly better than SwAV~\citep{caron2020unsupervised} and SCAN method~\citep{van2020scan} (based on MoCo~\citep{He2020MomentumCF} features) for image clustering. For this, similar to SCAN methods~\citep{van2020scan}, we picked random subsets of ImageNet categories, consisting of $50$, $100$ and $200$ classes. For this experiment, we were using validation images of ImageNet (50 images per category). The results show that DINO features could be used for image clustering with performance comparable with supervised ResNet-50 features.

\begin{table}
\caption{Clustering of random subsets of ImageNet classes.}
\label{tab:features}
\centering
\begin{tabular}{lccc}
\toprule
                                        & \multicolumn{3}{c}{Top-1 Accuracy, \%}                            \\ \cmidrule{2-4} 
                                        & \multicolumn{1}{c}{$50$ Classes} & \multicolumn{1}{c}{$100$ Classes} & \multicolumn{1}{c}{$200$ Classes} \\ \midrule
SCAN         & \multicolumn{1}{c}{$76.8$}       & \multicolumn{1}{c}{$68.9$}        & $58.1$                             \\ 
SwAV  & \multicolumn{1}{c}{$81.6 \pm 0.5$} & \multicolumn{1}{c}{$71.5 \pm 0.4$}  & $59.2 \pm 0.8$                       \\ 
Supervised & \multicolumn{1}{c}{$91.2 \pm 0.9$} & \multicolumn{1}{c}{$87.5 \pm 0.3$}  & $82.2 \pm 0.4$                       \\ \midrule
DINO & \multicolumn{1}{c}{$\mathbf{91.3 \pm 0.5}$} & \multicolumn{1}{c}{$\mathbf{88.0 \pm 0.2}$} & $\mathbf{83.1 \pm 0.4}$ \\ \bottomrule
\end{tabular}
\end{table}

\section{Saliency masks quality}
\label{sec:sal}
In this section, we study how \method works with different self-supervised and supervised saliency detectors. Overall, we observe that improving original object proposal masks is important for both category discovery and further iterative self-training. Also, we note that the proposed iterative self-training from filtered pseudo masks is effective for all of the studied choices of saliency detectors.
\subsection{Choice of unsupervised saliency masks detector}
\label{sec:different_sal}
First, we compare \method combined with self-supervised BasNet saliency detector~\citep{Qin_2019_CVPR}~(i.e., BasNet pretrained with DeepUSPS masks on MSRA-B dataset) with \method that is using recently proposed Spectral Decomposition saliency masks from DeepSpectral~\citep{melaskyriazi2022deep}. While original predictions from Spectral Decomposition are performing worse than unsupervised semantic segmentation proposed in DeepSpectral~(see the first row of the ~\Tab{tab:self_sup_sal}), using them in \method as objects proposal performs better than  DeepSpectral showing the importance of other \method components, such as the initial discovery of object categories from object proposals, not from clustered segments (as those are not always object-centric and could cover only parts of objects), filtering of the most uncertain pseudo-masks and several iterations of self-training on previously unseen data. This way, we are able to outperform DeepSpectral semantic segmentation on more than 5 mIoU points, while using only DINO features for both object proposal masks and object representation extraction.

Next, we additionally show how \method performs if we use original DeepUSPS masks~\citep{Nguyen_Neurips2019} as object proposals. As we discussed in the main text, DeepUSPS seems to be less robust than self-supervised BasNet on more complex OOD images from the PASCAL VOC dataset, that was not used in the training (see the first row of the ~\Tab{tab:self_sup_sal}). Additional self-training iterations are improving the quality of the original object proposals similarly, showing that \method can operate with originally lower quality mask proposals in case of successful object categories discovery. 

While the original DeepUSPS initialize its backbone weights from supervised pretraining~\citep{Nguyen_Neurips2019}, similar to other areas, it was recently shown that this is not necessary for DeepUSPS good performance~\citep{ryali2021learning}. They show that DeepUSPS$^2$~\citep{ryali2021learning} model that does not use any annotations during backbone pretraining still performs comparable to or better than the original DeepUSPS. Thus, similar to the original DeepUSPS saliency masks could be discovered from architecture where no labels were used. As DeepUSPS$^2$ implementation is not publically available and for better comparison with previous methods in our work, we were using self-supervised BasNet from MaskContrast~\citep{vangansbeke2020unsupervised}. 

\begin{table}[h]
\footnotesize
\renewcommand{\arraystretch}{1.2}
\centering
\caption{Choice of the unsupervised salient object detector. We compared \method performance with three different unsupervised saliency masks detectors: self-supervised BasNet model~\citep{Qin_2019_CVPR}, original DeepUSPS~\citep{Nguyen_Neurips2019} and spectral decomposition saliency masks from DeepSpectral~\citep{melaskyriazi2022deep}. All the models are evaluated with by IoU after Hungarian matching on the PASCAL 2012 \emph{val} set. }

\label{tab:self_sup_sal}
\begin{adjustbox}{max width=\textwidth}
  \setrow{\bfseries}
  \begin{tabular}{l@{\hskip1em}*{1}{c}{c}{c}@{\hskip1em}c<{\clearrow}}
  \toprule
   & Self-supervised BasNet & Spectral Decomposition & DeepUSPS 
     \\ \midrule
    \method (Iteration 1) &  47.6 & 43.8 & 45.5\\ 
    \method (Iteration 2) &	50.0 & 45.9 & 47.5 \\
  \bottomrule
  \end{tabular}
\end{adjustbox}
\end{table}

\subsection{Supervised saliency masks as localization prior}
\label{sec:sup_sal}
In addition to using fully self-supervised saliency masks as localization prior, we also consider using BasNet saliency detector trained with supervision on MSRA-B dataset as localization prior. Supervised training of saliency masks leads to even better masks, but drops the property of the method being fully unsupervised. \Tab{tab:sup_sal} shows that supervised saliency masks also improve the final results, as to be expected.

\begin{table}[h]
\footnotesize
\renewcommand{\arraystretch}{1.2}
\centering
\caption{Effect of object proposals from supervised saliency detector.}
\label{tab:sup_sal}
\begin{adjustbox}{max width=\textwidth}
  \setrow{\bfseries}
  \begin{tabular}{l@{\hskip1em}*{1}{c}{c}@{\hskip1em}c<{\clearrow}}
  \toprule
   & Unsupervised Saliency & Supervised Saliency 
     \\ \midrule
MaskContrast & 35.1	& 38.9 \\ \midrule
\method (Iteration 1) &  47.6 &  50.4 \\ 
\method (Iteration 2) &	50.0 &	52.3 \\
  \bottomrule
  \end{tabular}
\end{adjustbox}
\end{table}
\section{Extended Limitations and Future Work}
\subsection{Number of semantic categories bias}
\label{sec:n_objects}
As we are starting our iterative self-training from pseudo-masks restricted to one foreground semantic category per image, it is natural to study how well \method can work on more complex images where it is more than one semantic category per image by using discovered categories as an additional signal. Thus, we study \method performance in comparison with DeepSpectral~\citep{melaskyriazi2022deep} on two subsets of PASCAL VOC \emph{val}: the first one is the subset where it is only one foreground semantic category~(\texttt{subset 1}) and the second one is the subset of images with two or more foreground semantic categories~(\texttt{subset 2}). We compare several iterations of self-training with DeepSpectral performance using two metrics: first is the standard mIoU showing the overall quality of the predictions. In addition, we compared the precision of recognizing each group. For example for the first subset, it is equal to the percentage of predictions that contain only one foreground semantic category prediction). 

For each method, as expected,  we observe that the quality of the predictions on the \texttt{subset 2} is worse than on the \texttt{subset 1}. Iterations of self-training are improving \method performance, and allowing reaching better quality (measured by mIoU) on more complex images from \texttt{subset 2} than overall DeepSpectral predictions. While improving overall prediction quality, the second iteration of  \method shows bias towards predicting one foreground semantic category per image. This is potentially due to bias in the PASCAL VOC \texttt{trainaug} dataset itself. In contrast, DeepSpectral tends to have the opposite bias toward predicting more than one category per image (i.e., it has lower precision for the first task and higher precision for the second task). This could be due to the image features clustering task that is used by DeepSpectral for segment extraction. Interestingly none of the methods predicts well the number of foreground masks in both subsets. Thus determining the right number of semantic categories for each image is still a challenging problem for unsupervised semantic segmentation and an interesting direction for future work.

\begin{table}[h]
\caption{Number of semantic categories bias. Performance of studied methods on two subsets of PASCALVOC \emph{val} dataset.}
\label{tab:n_objects}
\centering
\hspace{0.2cm}
\begin{adjustbox}{max width=1.0\textwidth}

\begin{tabular}{l cccccccc}
\toprule
& \multicolumn{2}{c}{$1$ semantic category} & {} & \multicolumn{2}{c}{$>1$ semantic category }& {} & \multicolumn{2}{c}{overall}   \\
\cmidrule{2-3} \cmidrule{5-6} \cmidrule{8-9} 
 & mIoU & precision, \% && mIoU & precision, \%  && mIoU & precision, \% \\
\midrule
          DeepSpectral &  39.9 & 50.9 && 34.5	& 65.7     && 37.1 & 56.2 \\
\midrule
\method (Itteration 1) &  52.0 & 50.5   && 39.9	& 61.8  && 47.6 & 54.6 \\
\method (Itteration 2) &  55.2 & 67.7 && 41.2	& 46.5  && 50.0 & 60.1 \\

\bottomrule
\end{tabular}
\end{adjustbox}
\end{table}

\subsection{Failure Modes Analysis}
\label{sec:falure_modes}
\begin{figure}[t]
   \includegraphics[width=\linewidth]{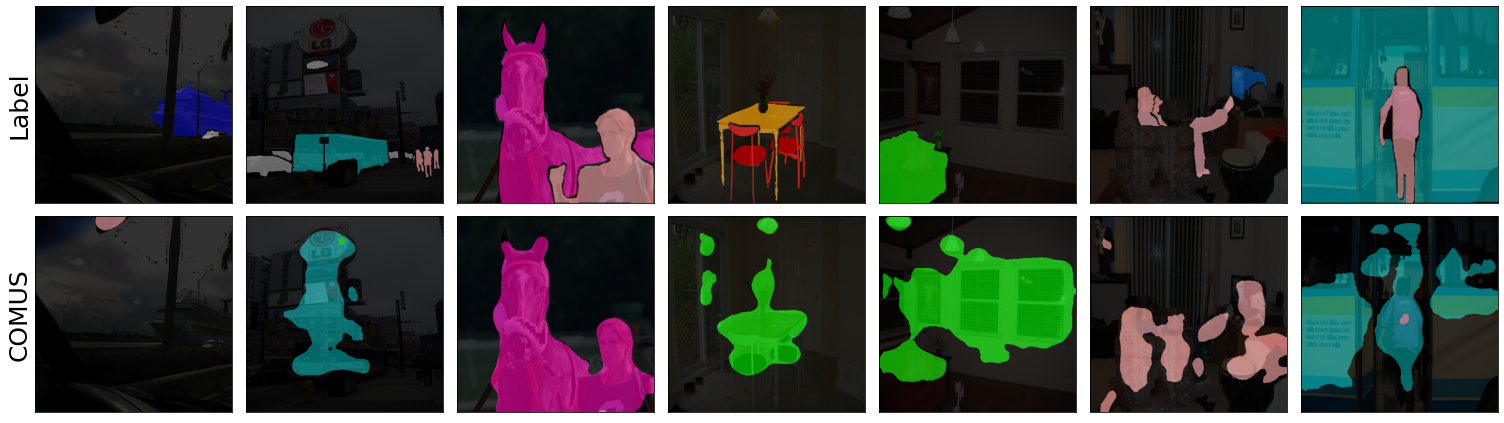}
   \caption{Several failure samples of the \method method on PASCAL VOC \textit{val} set. The failures  show the limitation and biases of our method, such as bias towards salient objects and misclassifications in multi-object images. }
   \label{fig:vis_falures}
\end{figure}

We present illustration of several failure cases in \Fig{fig:vis_falures}.  We observe that \method still failures to discover some categories, such as \texttt{table} category that is treated as a background by saliency method. Also, when an object fills all the background our method fails to fully recover from initial saliency mask assumption, that objects appears only in the foreground. Finally, as our method has only one category for the background, extending \method to additionally split backgrounds~(e.g., COCO-\emph{Stuff} semantic segmentation masks) is an interesting direction for future work.

\section{More detailed quantitative and qualitative results}
\subsection{PASCAL VOC}
In this subsection, we present additional analysis of COMUS performance on PASCAL VOC dataset. First, in \Tab{tab:iterative_detailed}, we show \method performance for each  PASCAL VOC category. \method performs better than MaskContrast and DeepSpectral on most of the categories. 
In addition to fully unsupervised semantic segmentation methods, we compare COMUS with recently proposed, weakly-supervised GroupViT method~\citep{Xu2022GroupViTSS}. GroupViT uses text descriptions as a weak supervision signal to group image regions into progressively larger arbitrary-shaped segments. While it does not require any pixel-level annotations, GroupViT still relies on large annotated datasets containing millions of image-text pairs. On average \method performance is worse than GroupViT performance, COMUS performs better on $9$ from $21$ categories while using no text annotations.  In addition, in \Fig{fig:images_pascal}, we visualize \method predictions on random images from PASCAL VOC dataset for two stages of \method as well as MaskContrast predictions. Finally, for exploring Interactive Demo that visualizes clustering of the whole PASCAL VOC \emph{val} set, visit \method project website: \url{https://sites.google.com/view/comuspaper/home}.
\begin{table}[h]
\footnotesize
\renewcommand{\arraystretch}{1.2}
\centering
\caption{More detailed comparison to prior art and iterative improvement via  self-training (evaluated by IoU after Hungarian matching) on the PASCAL 2012 \emph{val} set. Our method results are averaged over 5 runs.}
\label{tab:iterative_detailed}
\begin{adjustbox}{max width=1.0\textwidth}
  \setrow{\bfseries}
  \begin{tabular}{r@{\hskip1em}*{21}{c}@{\hskip1em}c<{\clearrow}}
  \toprule
  \textbf{Method} & \tableIcons & \textbf{mIoU}
     \\ \midrule
  \emph{GroupViT (text supervision)}~\citep{Xu2022GroupViTSS} & 80.6 & 38.1 & 31.4 & 51.6 & 32.7 & 63.5 & 78.8 & 65.1 & 79.2 & 18.8 & 73.4 & 31.6 & 76.4 & 59.4 & 55.3 & 44.0 & 40.9 & 66.6 & 31.5 & 49.5 & 29.7 & 52.3 \\
  \midrule
  MaskContrast~\citep{vangansbeke2020unsupervised} & 84.4 & 39.1 & 20.0 & 59.5 & 34.2 & 38.1 & 57.8 & 60.7 & 46.9 & 0.29 & 0.42 & 3.51 & 28.6 & 39.6 & 54.7 & 23.2 & 0.00 & 40.0 & 14.9 & 54.0 & 37.7 & 35.1 \\
DeepSpectral~\citep{melaskyriazi2022deep} & 82.1& 46.1& 0.0& 72.6& 31.9& 9.1& 77.3& 66.1& 77.5& 0.1& 43.4& 25.9& 40.6& 62.6& 36.9& 28.0& 2.5& 1.1& 10.8& 63.9& 0.0& 37.1 \\
  \midrule
  \setrow{\bfseries}
  Pseudo-masks (Iteration 0) & 82.8& 48.1& 19.6& 59.3& 49.7& 55.6& 63.8& 52.7& 53.2& 0.1& 58.3& 0.0& 37.2& 54.7& 50.9& 36.2& 26.8& 66.8& 17.6& 52.3& 33.9& 43.8 $\pm$ 0.1\\
  \method (Iteration 1) & 85.9& 51.0& 21.4& 49.4& 52.5& 61.0& 71.0& 61.6& 65.4& 0.0& 47.1& 16.7& 59.9& 48.9& 56.9& 49.6& 33.9& 58.9& 16.1& 56.5& 36.4& 47.6 $\pm$ 0.4\\
  \method  & 86.3& 54.8& 21.9& 53.5& 55.3& 64.5& 75.2& 61.8& 68.6& 0.0& 49.0& 18.2& 64.1& 52.2& 58.3& 52.4& 36.8& 57.7& 16.8& 63.5& 38.7& \textbf{50.0 $\pm$ 0.4}\\
  
  \bottomrule
  \end{tabular}
\end{adjustbox}
\end{table}

\subsection{COCO}
\label{sec:coco_details}
\begin{figure}[ht]
    \centering
    \includegraphics[width=0.37\linewidth]{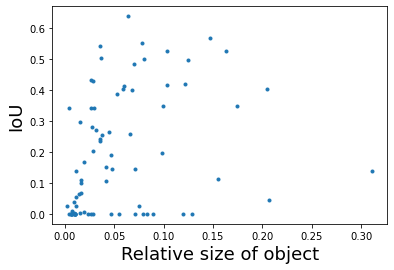}
    \caption{Connection between mean relative size of objects and IoU for each COCO category. The Spearman's rank correlation between relative size and IoU is equal to $0.43$.  }
    \label{fig:coco_size_iou}
\end{figure}

\begin{figure}[ht]
    \centering
    \includegraphics[width=0.75\linewidth]{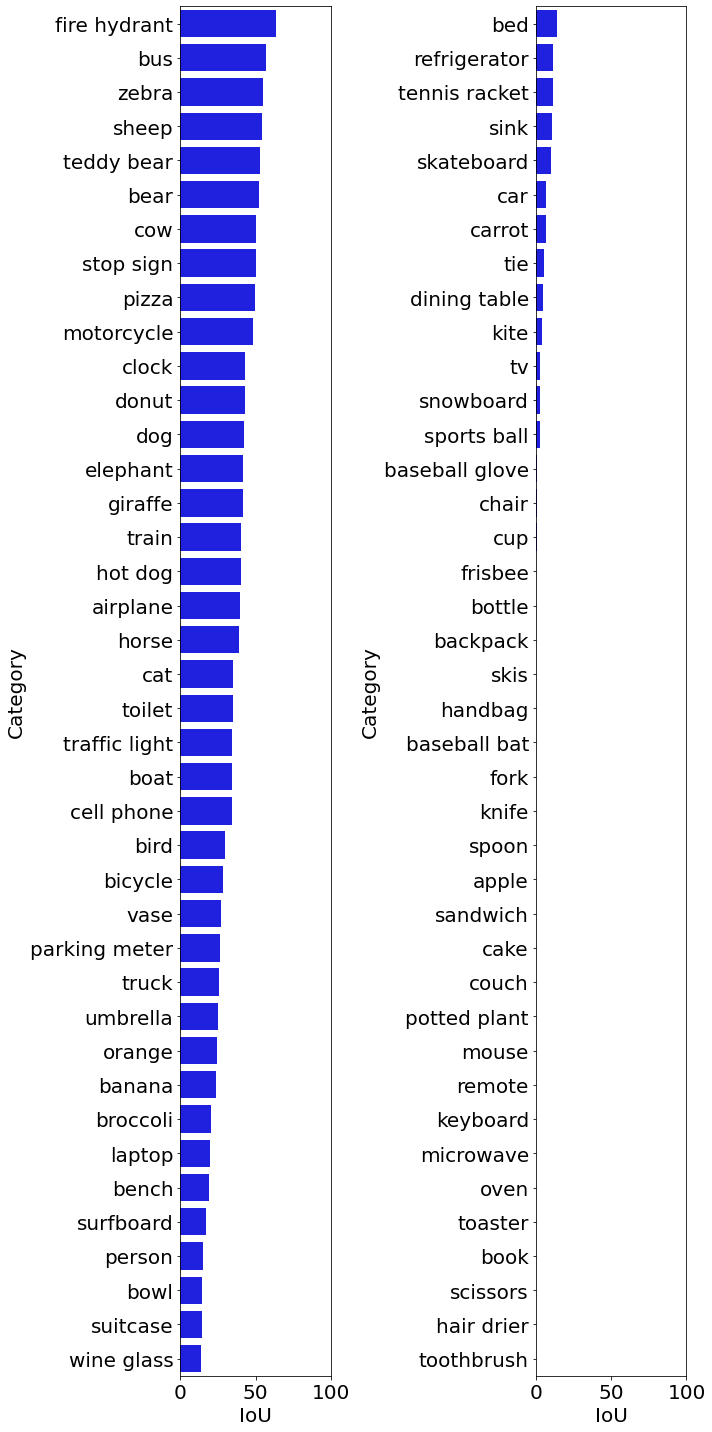}
    \caption{IoU for COCO categories after Hungarian matching of the cluster IDs to ground-truth categories. }
    \label{fig:coco_details}
\end{figure}

\begin{figure}
\begin{minipage}[t]{0.6\textwidth}
   \centering
   \includegraphics[width=\linewidth]{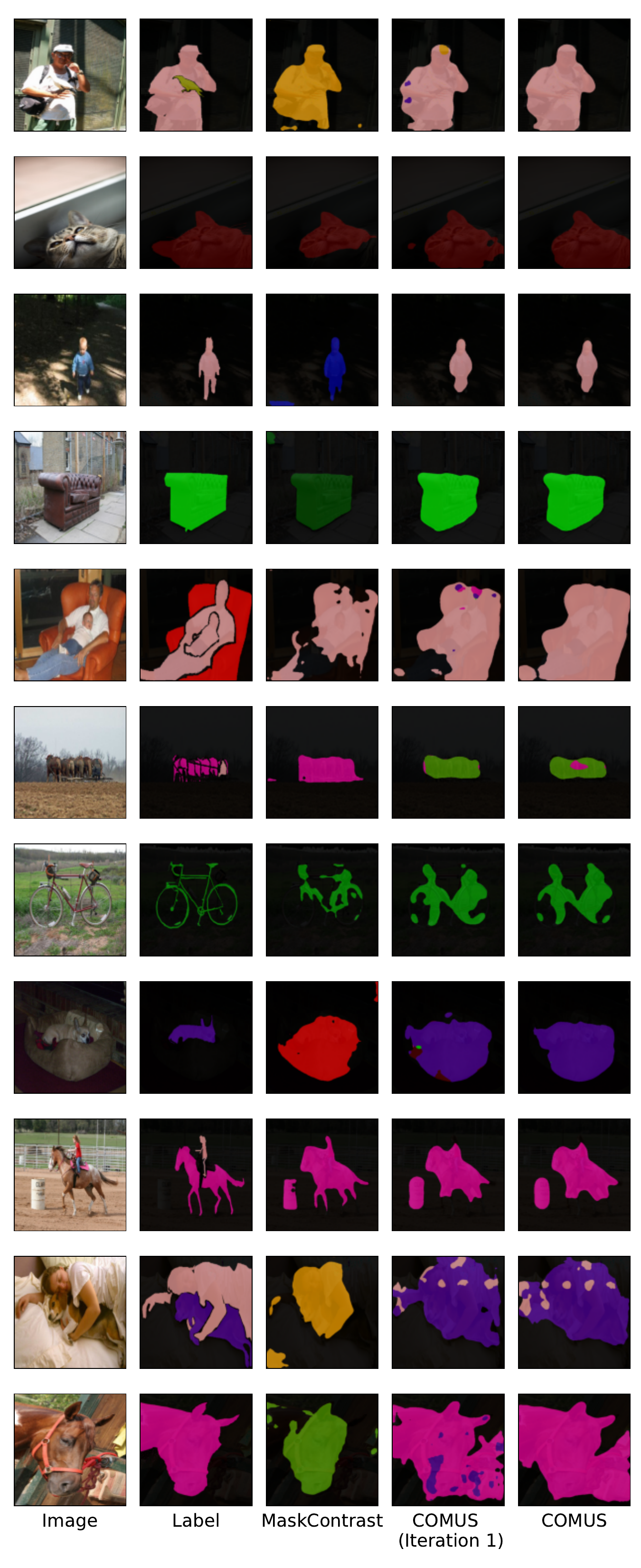}
   \captionof{figure}{\method and MaskContrast predictions on random images from PASCAL VOC \textit{val} set. }
   \label{fig:images_pascal}
\end{minipage}
\hfill
\begin{minipage}[t]{0.36\textwidth}

    \centering
   \includegraphics[width=\linewidth]{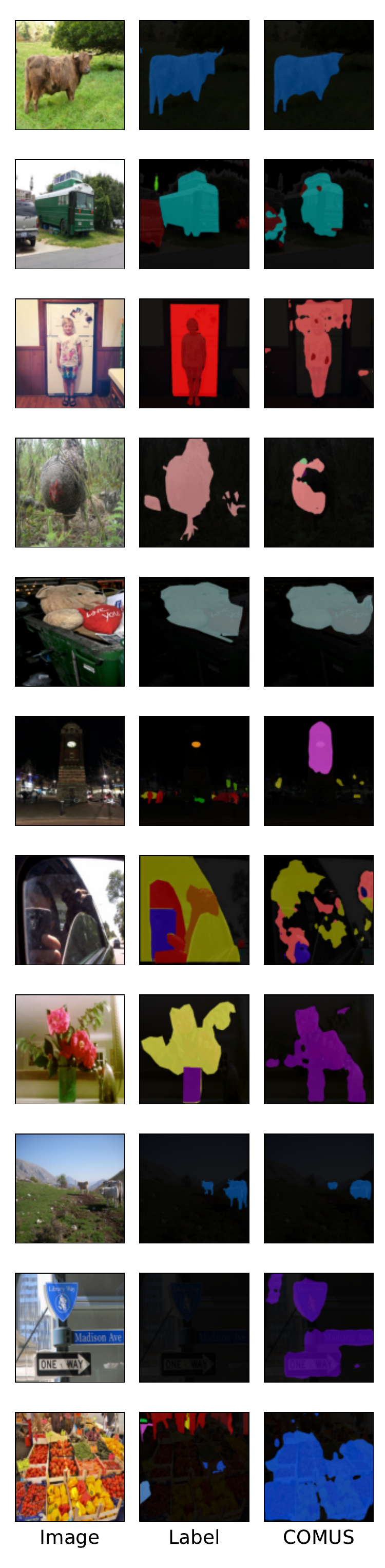}
   \captionof{figure}{\method predictions on random images from COCO \textit{val} set. }
   \label{fig:images_coco}
\end{minipage}
\end{figure}

In this subsection, we present additional analysis of \method performance on COCO dataset. \Fig{fig:coco_details} shows the performance of \method on COCO for each of 80 COCO categories. We note that \method achieves better performance for animal and vehicle categories, as well as salient object categories such as \texttt{stop sign} and \texttt{traffic light} categories. We also observe that most of the undiscovered categories have small relative object's size (e.g., \texttt{spoon}, \texttt{remote} and \texttt{mouse}). For additional analysis of the connection between relative object's size and \method performance, see \Fig{fig:coco_size_iou}. In addition, in \Fig{fig:images_coco}, we visualize \method predictions on random images from PASCAL VOC dataset for two stages of \method as well as MaskContrast predictions. 

\begin{figure}
   \includegraphics[width=\linewidth]{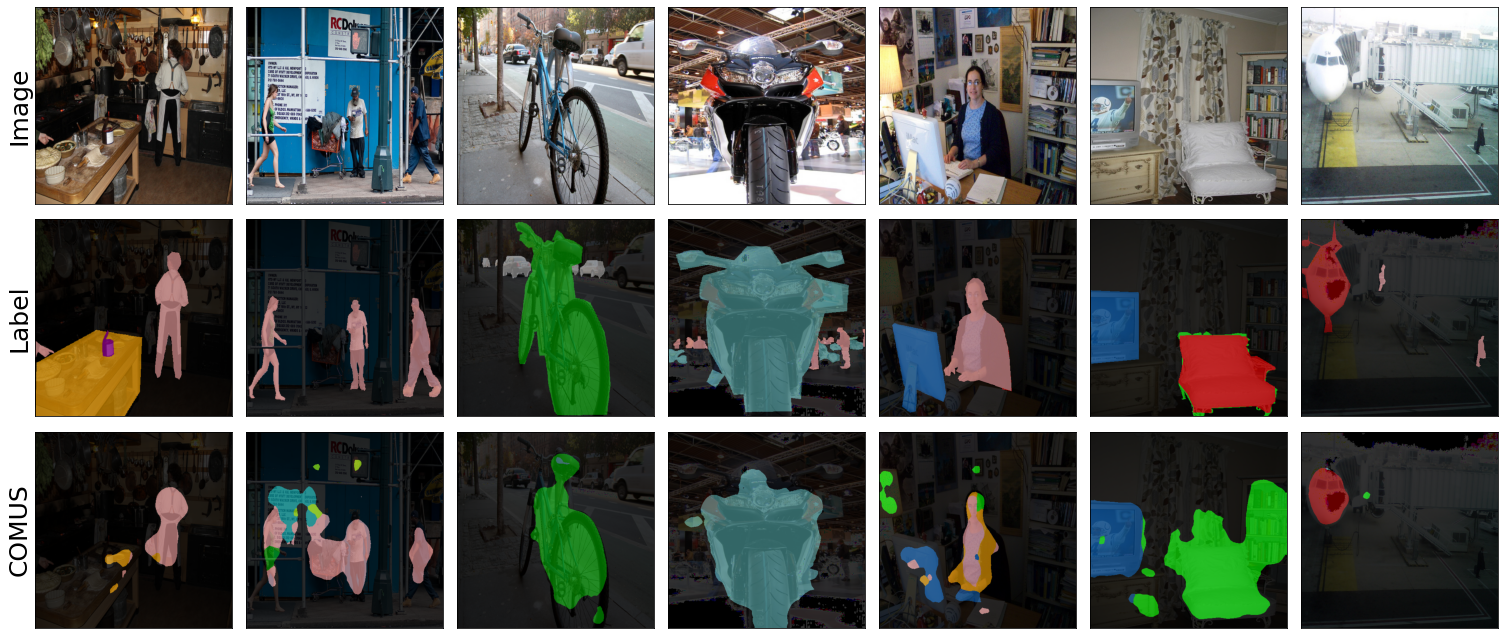}
   \caption{Predictions of the \method method trained on PASCAL VOC on COCO \textit{val} set. We notice that the predictions from models trained on PASCAL VOC transfer reasonably well to COCO. }
   \label{fig:transfer}
\end{figure}

\section{Implementation details}
\label{sec:parameters}
\subsection{Spectral Clustering}
We use spectral clustering implementation from sklearn\footnote{https://scikit-learn.org/stable/modules/generated/sklearn.cluster.SpectralClustering.html}. In particular, an affinity matrix is obtained from the nearest neighbors graph. Number of eigenvectors and number of clusters is the same as number of GT categories. We refer to \Tab{tab:sc-hyperparams} for the parameters of spectral clustering.
\begin{table}[ht]
    \centering
    \caption{Spectral clustering parameters for \method on PASCAL VOC and MS COCO datasets.}
\label{tab:sc-hyperparams}
    \begin{tabular}{ccc}
    \toprule
    \textbf{Hyper-parameter} & \textbf{PASCAL VOC} & \textbf{MS COCO} \\
    \midrule
    Number of clusters & 20 & 80  \\
    Number of components & 20 & 80 \\
    Affinity & \texttt{nearest neightbors} & \texttt{nearest neightbors} \\
    Number of neighbors & 30 & 30 \\ 
    \bottomrule
    \end{tabular}
\end{table}

\subsection{Self-Training}
During self-training, DeepLabv3~\citep{Chen2018DeepLabSI} with standard cross-entropy loss is chosen to make training set up as comparable to previous research (e.g. to MaskContrast~\citep{vangansbeke2020unsupervised}) as possible. We use \texttt{CropResize} and \texttt{HorizontalFlip} as data augmentation methods. For PASCAL VOC, we perform two iterations of self-training on \textit{train} and \textit{trainaug} sets. For COCO dataset, we perform one iteration of self-training on \textit{train} set. See \Tab{tab:st-hyperparams} for the parameters of the self-training.

\subsection{Computational requirements}
\label{sec:comp_requirements}
Similar to other unsupervised segmentation methods~\citep{vangansbeke2020unsupervised, melaskyriazi2022deep, Hamilton2022Stego}, the most expensive part of our pipeline, is the training of self-supervised representation learning method. In particular, DINO with Vision Transformers training takes 3 days on two 8-GPU servers and is comparable with other self-supervised representation learning methods. However, learned features could be transferred without further fine-tuning on new natural data images, such as scenes from PASCAL VOC and MS COCO.  

The other parts of our method are much faster to train: DeepUSPS could be trained 30 hours of computation time on old four Geforce Titan X~\citep{Nguyen_Neurips2019}, while BasNet could be trained with four GTX 1080ti GPU (with 11GB memory) in around 30 hours~\citep{Qin_2019_CVPR}, while the inference for 256×256 image only takes 0.040s (25 fps). For more details, we refer the reader to the main papers for these methods.

Finally, the main parts of our method that do require training on new data are object proposals clustering and self-training iterations. Spectral Clustering complexity depends on the sample size. For the full COCO \texttt{train} dataset Spectral Clustering with \texttt{amg} solver took 20 minutes on 96 core node. Thus, for large datasets, it is recommendable to use its subset for the initial discovery of object categories and then use self-training on the full dataset.
For semantic segmentation self-training, we used the standard in supervised semantic segmentation set up for training DeepLabv3 architecture. While all the models could be trained on a single GPU, for convenience we perform all the experiments on one node with 4 NVIDIA T4 GPUs, where 2 iterations of self-training took around one hour.

\begin{table}[t]
    \centering
    \caption{Self-training parameters for \method on PASCAL VOC and MS COCO datasets.}
    \label{tab:st-hyperparams}
    \begin{tabular}{ccc}
    \toprule
    \textbf{Hyper-parameter} & \textbf{PASCAL VOC} & \textbf{MS COCO} \\
    \midrule
    Optimizer & Adam with default settings & Adam with default settings \\
    Learning rate & $0.00006$ & $0.00006$  \\
    Batch size & $56$ & $56$ \\ 
    Input size & $512$ & $256$  \\
    Crop scales & $[0.5, 2]$ & $[0.2, 1.0]$  \\
    Number of iterations & $2$ & $1$ \\  
    Number of epochs. Iteration 1 & $10$ & $1$ \\    
    Number of epochs. Iteration 2 & $5$ & - \\     
    \bottomrule
    \end{tabular}
\end{table}

\section{Datasets (directly or indirectly) used in the paper}
\label{sec:datasets}
\paragraph{PASCAL VOC:} The PASCAL Visual Object Classes (VOC) project~\citep{Everingham15} provides different datasets / challenges for the years from 2005 to 2012. We apply our proposed method to the datasets from 2012 and 2007, which come with semantic segmentation masks. All datasets and detailed descriptions are available on the PASCAL VOC homepage (\url{http://host.robots.ox.ac.uk/pascal/VOC/index.html}).

\paragraph{MS COCO:} We also apply our method to the Microsoft (MS) COCO dataset~\citep{Lin2014MicrosoftCC}. The dataset and informations are available on \url{https://cocodataset.org/#home}.

\paragraph{ImageNet:} For feature extraction, we use vision transformers pretrained with the self-supervised (no labels!) DINO method \citep{caronEmergingPropertiesSelfSupervised2021} on ImageNet~\citep{imagenet_cvpr09}. The pretrained checkpoint can be found on \url{https://github.com/facebookresearch/dino}. Informations about ImageNet are provided on \url{https://image-net.org/}. 

\paragraph{MSRA-B:} For computing saliency mask, we use BasNet~\citep{Qin_2019_CVPR} pretrained on pseudo-labels generated with the unsupervised DeepUSPS~\citep{Nguyen_Neurips2019} outputs on the MSRA-B dataset~\citep{WangDRFI2017}. The pretrained checkpoint can be found on 
\url{https://github.com/wvangansbeke/Unsupervised-Semantic-Segmentation/tree/main/saliency}.  The dataset and informations about it 
are available on \url{https://mmcheng.net/msra10k/}.
\end{document}